\documentclass[10pt,twocolumn,letterpaper]{article}
\usepackage{iccv}
\usepackage{times}
\usepackage{epsfig}
\usepackage{graphicx}
\usepackage{amsmath}
\usepackage{amssymb}
\usepackage{subfigure}

\usepackage{booktabs}
\usepackage[permil]{overpic}
\usepackage{color}
\usepackage[dvipsnames]{xcolor}
\usepackage{algpseudocode}
\usepackage{wrapfig}
\usepackage{algorithm}
\usepackage{bm}
\usepackage{listings}
\usepackage{diagbox}
\usepackage{mathtools}
\usepackage{multirow}
\usepackage{listings}
\usepackage[font=small]{caption}
\usepackage{bbding}
\usepackage{pifont}
\usepackage{wasysym}

\usepackage[accsupp]{axessibility} 

\usepackage{etoolbox}
\makeatletter
\AfterEndEnvironment{algorithm}{\let\@algcomment\relax}
\AtEndEnvironment{algorithm}{\kern2pt\hrule\relax\vskip3pt\@algcomment}
\let\@algcomment\relax
\newcommand\algcomment[1]{\def\@algcomment{\footnotesize#1}}
\renewcommand\fs@ruled{\def\@fs@cfont{\bfseries}\let\@fs@capt\floatc@ruled
  \def\@fs@pre{\hrule height.8pt depth0pt \kern2pt}%
  \def\@fs@post{}%
  \def\@fs@mid{\kern2pt\hrule\kern2pt}%
  \let\@fs@iftopcapt\iftrue}
\makeatother

\definecolor{T1}{RGB}{239,146,181}
\definecolor{T2}{RGB}{192,208,157}
\definecolor{Stu}{RGB}{172,204,214}

\usepackage[breaklinks=true,bookmarks=false]{hyperref}
\iccvfinalcopy

\def\iccvPaperID{11317} 

\ificcvfinal\fi

\begin{document}

\title{Periodically Exchange Teacher-Student for Source-Free Object Detection}

\author{Qipeng Liu, Luojun Lin\footnotemark[1], Zhifeng Shen, Zhifeng Yang \\
{\normalsize College of Computer and Data Science, Fuzhou University}\\
{\tt\small lqpwiki@gmail.com, linluojun2009@126.com, \{shen\_zhifeng, yzf2001\}@outlook.com}
}

\maketitle
\ificcvfinal\thispagestyle{empty}\fi
\renewcommand{\thefootnote}{\fnsymbol{footnote}}
\footnotetext[1]{Corresponding author}

\begin{abstract}
Source-free object detection (SFOD) aims to adapt the source detector to unlabeled target domain data in the absence of source domain data. Most SFOD methods follow the same self-training paradigm using mean-teacher (MT) framework where the student model is guided by only one single teacher model. However, such paradigm can easily fall into a training instability problem that when the teacher model collapses uncontrollably due to the domain shift, the student model also suffers drastic performance degradation. To address this issue, we propose the Periodically Exchange Teacher-Student (PETS) method, a simple yet novel approach that introduces a multiple-teacher framework consisting of a static teacher, a dynamic teacher, and a student model. During the training phase, we periodically exchange the weights between the static teacher and the student model. Then, we update the dynamic teacher using the moving average of the student model that has already been exchanged by the static teacher. In this way, the dynamic teacher can integrate knowledge from past periods, effectively reducing error accumulation and enabling a more stable training process within the MT-based framework. Further, we develop a consensus mechanism to merge the predictions of two teacher models to provide higher-quality pseudo labels for student model. Extensive experiments on multiple SFOD benchmarks show that the proposed method achieves state-of-the-art performance compared with other related methods, demonstrating the effectiveness and superiority of our method on SFOD task.
\end{abstract}

\section{Introduction}
Object detection has achieved significant progress with rapid development of dataset scale and computation capability~\cite{faster_r-cnn,yolov6, diffusiondet}. However, these detectors are typically trained under an i.i.d assumption that the train and test data are independently and identically distributed, which does not always hold in real-world due to the existence of domain shift between the train and test data. This can cause significant performance degradation when applying a model well-trained on source domain (train data) to the target domain (test data). Unsupervised domain adaptation (UDA), a recent research hotspot, can resolve this dilemma by enabling the model to adapt effectively to the target domain. This is achieved through joint training, leveraging both labeled source domain data and unlabeled target domain data to enhance the model's performance in the target domain.

\begin{figure}[t]
\begin{center}
\includegraphics[width=1.0\linewidth]{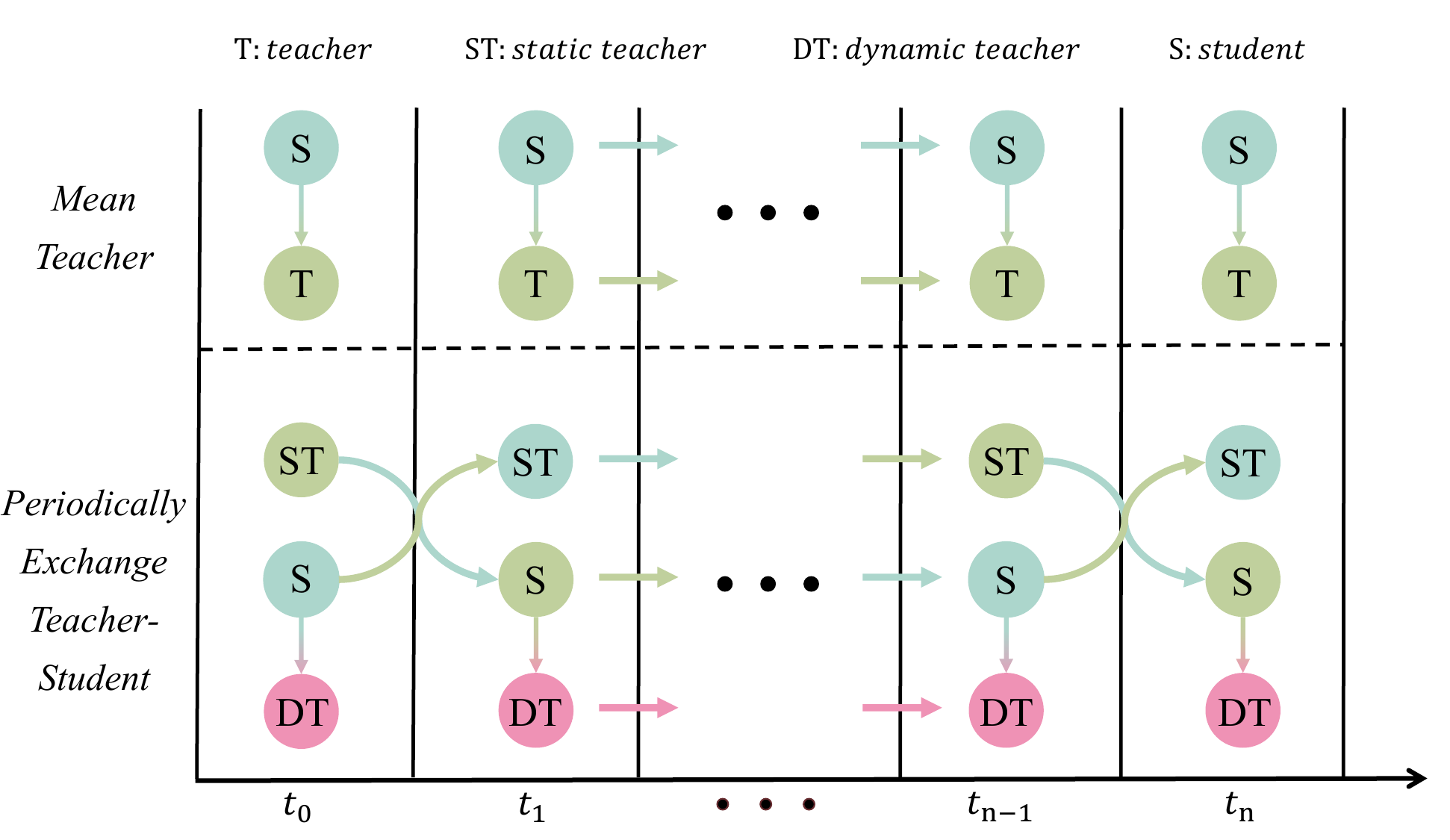}
\end{center}
   \vspace{-0.3cm}
   \caption{The training paradigms of the \emph{mean-teacher} and the proposed \emph{periodically exchange teacher-student} method. \textbf{T} and \textbf{S} denote the teacher model and student model, respectively. \textbf{ST} represents the static teacher with fixed weights in each period, and \textbf{DT} is the dynamic teacher updated by the EMA of the student models. $t_{i}$ represents the $i$-th period in whole training stage.}
    \label{fig:1}
\end{figure}

There are many UDA methods developed to address domain shift in image classification tasks~\cite{pseudo_shot, KUDA,adversarial_adda,tvt}. However, these methods cannot meet the growing demand for data privacy protection. Moreover, directly applying these UDA methods to object detection tasks cannot achieve satisfactory performance. In light of the above considerations, source-free object detection (SFOD) has rapidly emerged as an urgent task to attract the attention of researchers. The purpose of SFOD is to achieve effective adaptation of a detector, originally trained on a labeled source domain, to the unlabeled target domain, without accessing any source data during adaptation. Compared with source-free image classification, SFOD is a more challenging task that not only requires regression, i.e., locating the bounding box of each object, but also involves classification, i.e., identifying the associated class of each object in diversely-scaled images.

\begin{figure}[htbp]
\begin{minipage}[t]{0.325\linewidth}
\centering
\includegraphics[width=1.0\linewidth]{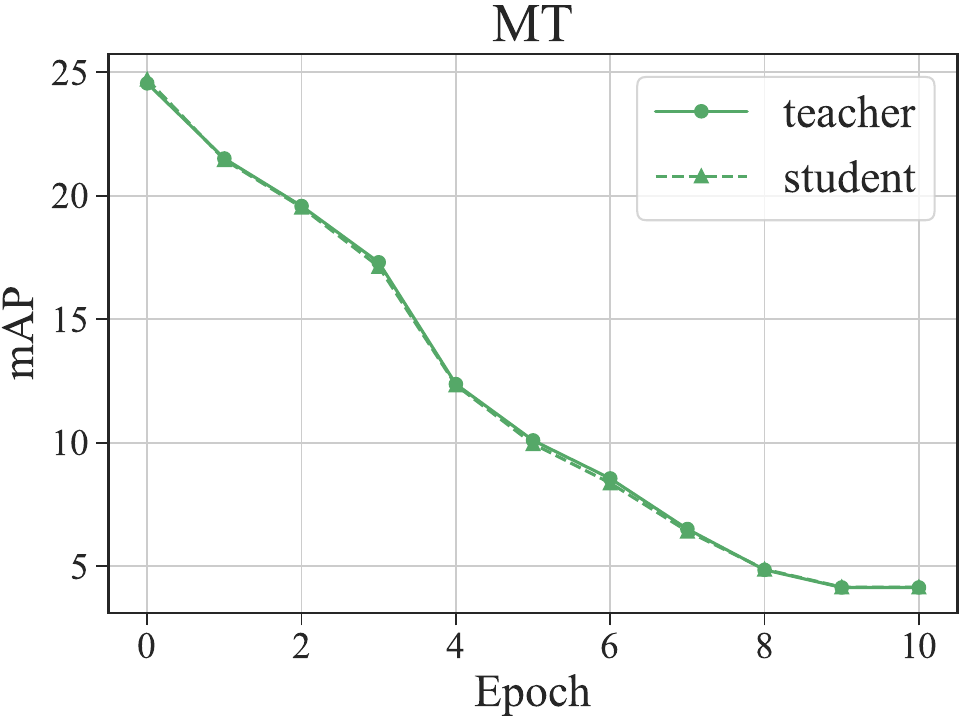}
\fontsize{7pt}{\baselineskip}\selectfont (a) EMA weight = 0.99
\end{minipage}
\begin{minipage}[t]{0.325\linewidth}
\centering
\includegraphics[width=1.0\linewidth]{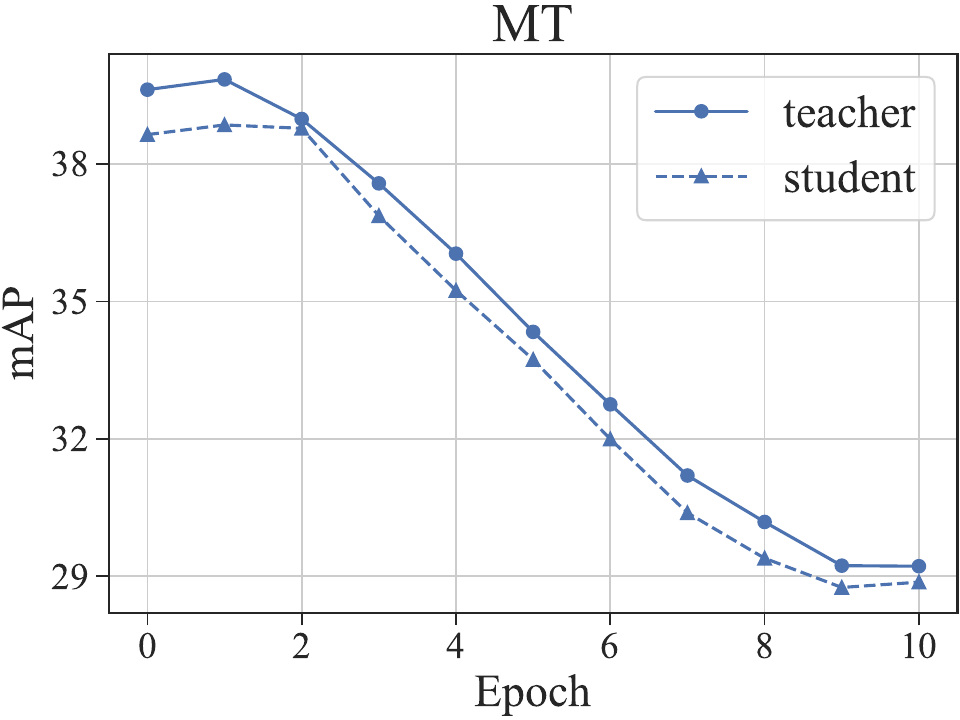}
\fontsize{7pt}{\baselineskip}\selectfont (b) EMA weight = 0.999
\end{minipage}
\begin{minipage}[t]{0.325\linewidth}
\centering
\includegraphics[width=1.0\linewidth]{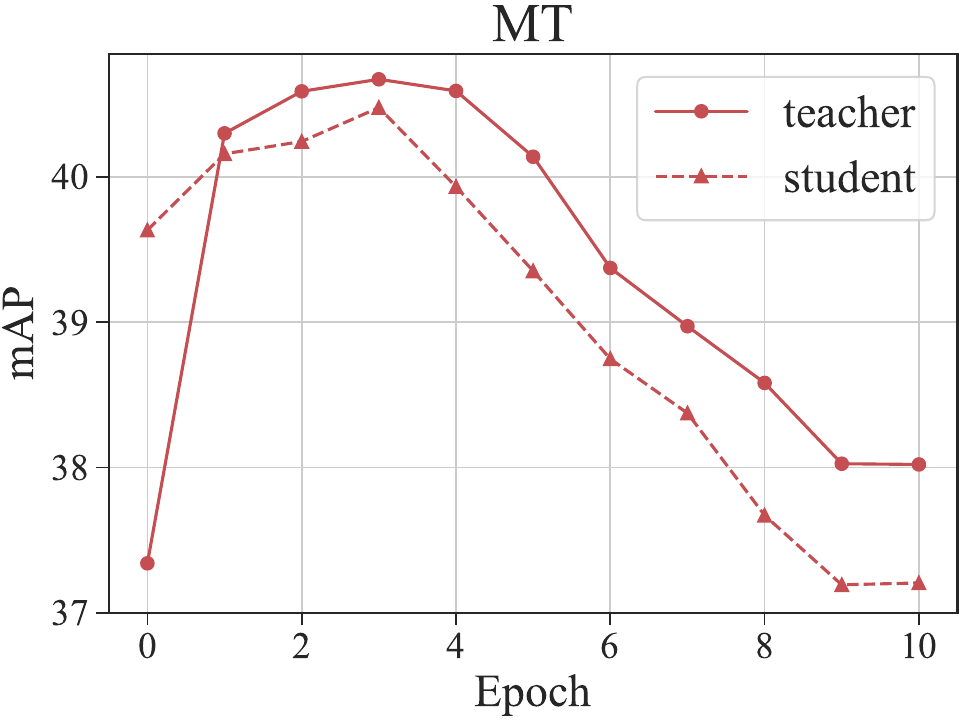}
\fontsize{7pt}{\baselineskip}\selectfont (c) EMA weight = 0.9996
\end{minipage}

\begin{minipage}[t]{0.33\linewidth}
\centering
\vspace{0.2cm}
\includegraphics[width=1.0\linewidth]{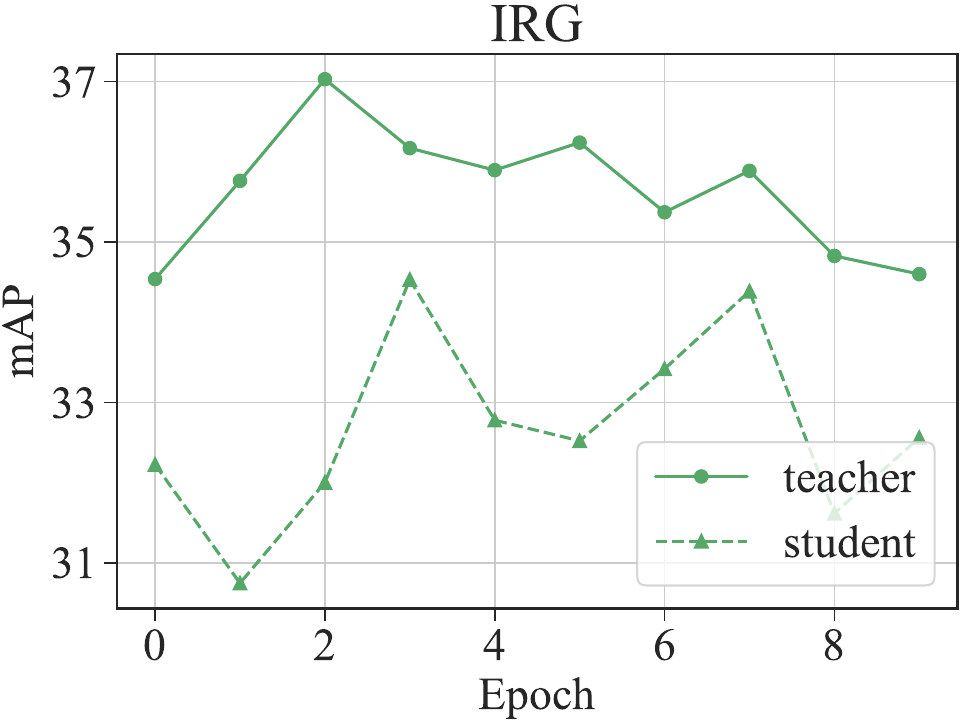} \\
\fontsize{7pt}{\baselineskip}\selectfont (d) EMA stepsize = 600
\end{minipage}%
\begin{minipage}[t]{0.33\linewidth}
\centering
\vspace{0.2cm}
\includegraphics[width=1.0\linewidth]{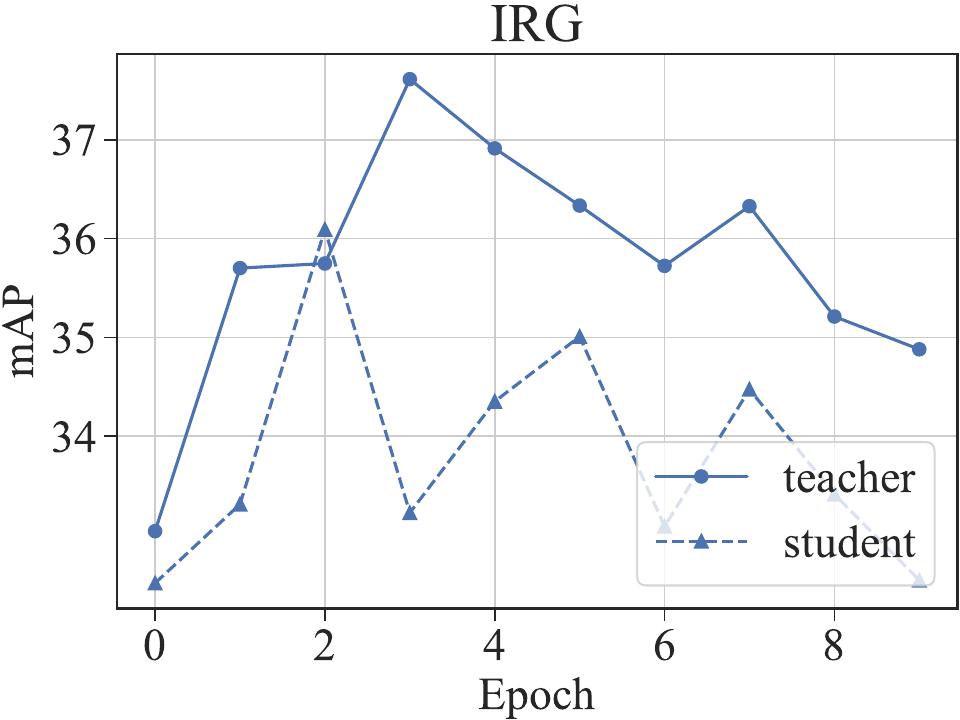}
\fontsize{7pt}{\baselineskip}\selectfont (e) EMA stepsize = 1500
\end{minipage}
\begin{minipage}[t]{0.33\linewidth}
\centering
\vspace{0.2cm}
\includegraphics[width=1.0\linewidth]{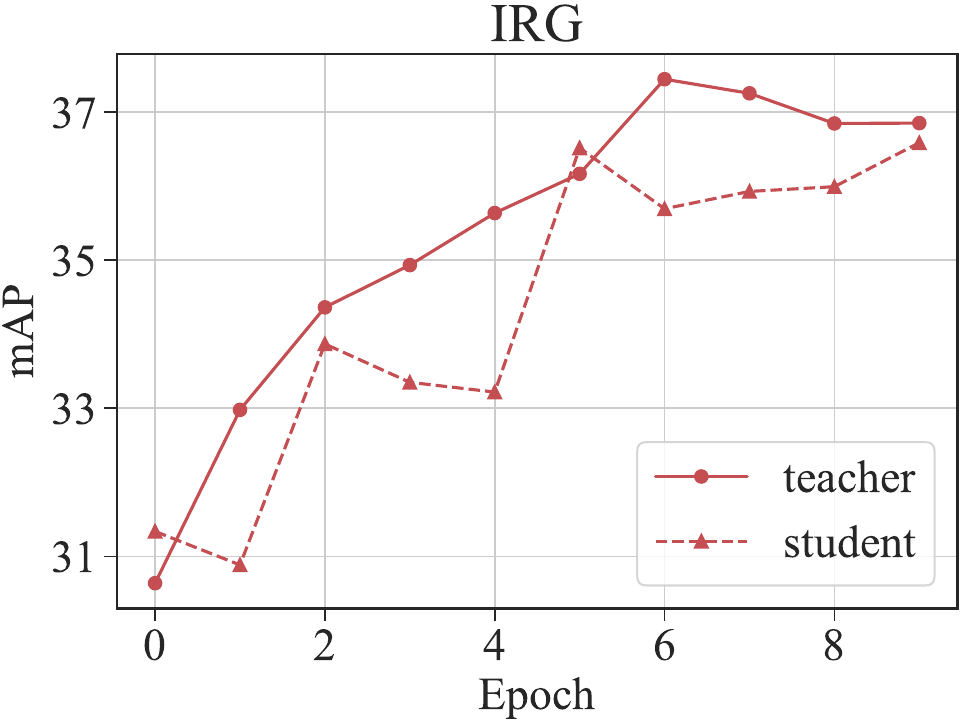}
\fontsize{7pt}{\baselineskip}\selectfont (f) EMA stepsize = 3000
\end{minipage}
\vspace{-0.2cm}
\caption{The training curves of different SFOD methods (i.e. conventional MT and IRG~\cite{IRG}) with different EMA hyper-parameters on C2F benchmark~\cite{cityscapes_foggy}. These methods show a consistent phenomenon: when the performance of the teacher model crashes, the student model always follows the downward trend of the teacher model even with different EMA weights or stepsizes.}
\label{fig:slow}
\vspace{-0.2cm}
\end{figure} 

Most of the existing SFOD studies~\cite{LODS, SOAP, SFOD,AASFOD,MoTE} are based on self-training paradigm using a mean-teacher (MT) framework~\cite{mean_teacher} along with other improved UDA techniques. These MT-based methods involves using a single teacher model to guide the student model, where the teacher model is an exponential moving average (EMA) of the student model at different time steps, and the student model is updated based on the pseudo labels provided by the teacher model. The MT framework assumes that \emph{the teacher model can be improved continuously as the training progresses, and the student model can gradually approach the performance of the teacher model}. However, since the source-pretrained model introduces inherent biases when applied to the target domain, the teacher model, as an EMA of the student model inherited from the source-pretrained model, is susceptible to accumulating errors from the student model. This error accumulation leads to a concerning issue of \emph{training instability} for the teacher model, thereby making the initial assumption no longer holds true. That is, when the single teacher model makes mistakes, the student model tends to replicate the errors without any correction measures. It finally leads to uncontrollable degradation of the detection performance for MT-based SFOD methods.

In order to mitigate the training instability problem, a natural solution involves adjusting the EMA hyper-parameters to encourage a more gradual and stable evolution of the teacher model. For example, the recent works \cite{IRG, AASFOD} have explored the strategy of employing a larger EMA update stepsize, with the aim of slowing down the updating process of the teacher model. Another line of exploration in this direction involves assigning a higher EMA weight to the historical teacher model, amplifying the influence of the past iterations and consequently reducing the updating rate of the teacher model. However, these efforts have yielded limited success. As shown in Figure \ref{fig:slow}, the efforts to enhance the EMA weights or increase the EMA update stepsize do not completely resolve the issue of training instability problem within the MT-based frameworks. Besides, it is inconvenient to search for an optimal EMA hyper-parameter to properly update the teacher model.

In this paper, we aim to address the instability problem and thus propose a simple yet novel \emph{Periodically Exchange Teacher-Student (PETS)} method to improve the self-training paradigm of the MT framework. As shown in Figure \ref{fig:1}, our method is a multiple-teacher framework consisting of a static teacher model, a dynamic teacher model, and a student model. Unlike the previous methods that keep the roles of student and teacher unchanged throughout the training, we periodically exchange the positions between the student model and the static teacher model. Then, the static teacher model freezes its weights until the next exchanging period; while the student model is trained using the supervision signals provided by the two teacher models, and the dynamic teacher model is updated by an EMA of the student per iteration within each period. In this way, the dynamic teacher implicitly reduces error accumulation to improve its performance. Moreover, the exchange between the static teacher and student helps to prevent a rapid decrease in the lower bound of the student model, ultimately improving the robustness of whole models in our method. Besides, we also propose a consensus mechanism to merge the predictions from the static and dynamic teachers, which can provide higher-quality pseudo labels to supervise the student model. 

Our method is evaluated on four SFOD benchmarks. The experimental results show that our method achieves competitive results compared with existing SFOD methods, and demonstrate its effectiveness to solve the instability problem of current MT-based frameworks. The main contributions of our method are summarized as follows: 
\begin{itemize}
\vspace{-0.1cm}
\setlength{\itemsep}{1pt}
\item We highlight the \emph{training instability} issue within the MT framework, where the errors from the teacher model can be replicated by the student model without correction measures. This will result in an uncontrollable degradation of detection performance in MT-based SFOD methods.
\item We propose a simple yet novel \emph{Periodically Exchange Teacher-Student (PETS)} method to address the training instability issue for MT framework. Our method consists of a static teacher, a dynamic teacher and a student model. At the end of each period of training, we exchange the weights between the student and the static teacher to reduce error accumulation. Within each period, we train the student model through the two teacher models, and update the dynamic teacher with an EMA of the student model per iteration. 
\item We design a consensus mechanism to integrate the predictions from the static teacher and the dynamic teacher models. It integrates knowledge from historical iterations to prevent catastrophic forgetting, which can achieve higher-quality pseudo labels to supervise the student model.
\item Extensive experiments on multiple SFOD benchmarks show that the proposed method achieves state-of-the-art performance compared with other related methods, demonstrating the effectiveness and superiority of our method on SFOD task.
\end{itemize}

\section{Ralated Works}
\subsection{Unsupervised Domain Adaptation}
Unsupervised Domain Adaptation (UDA) aims to transfer knowledge from a source domain with labeled data to a target domain without labeled data. The current UDA methods can be roughly categorized into three types: domain translation, adversarial learning and pseudo labeling. The domain translation methods aim to transform a target image into a source-like image by using statistic information in the model~\cite{bn_gen1,bn_gen3} or employing a translation network~\cite{DM,UMT, afan}. Adversarial learning is also frequently adopted in UDA tasks by employing a domain discriminator~\cite{adversarial_dann} or designing adversarial loss functions, in order to narrow the gap between source and target domains in feature space~\cite{adversial_6, adversarial_adda, adversial_5, meng2022slimmable}. Unlike previous methods, pseudo labeling, as one of the most popular self-training paradigms~\cite{chen2020unsupervised}, has been an effective approach for UDA, which is mainly constructed based on the mean-teacher (MT) framework~\cite{mean_teacher} that exploits the pseudo labels provided by the teacher model to supervise the student model. Most pseudo labeling methods concentrate on designing interaction manners between the student and teacher models~\cite{NL,ssnll,AML}. In this paper, we concentrate on source-free object detection and try to improve self-training paradigm for MT-based SFOD framework.

\subsection{Source-Free Object Detection}
Several UDA approaches have been applied to \emph{Unsupervised Domain Adaptive Object Detection (UDAOD)}, which can also be categorized into adversarial learning~\cite{DA-Faster,SW,MAF}, domain translation~\cite{PDA,DM} and pseudo labeling~\cite{mtor,UMT}. Given that these methods have been introduced briefly in previous section, we only discuss the final one since our work is constructed on the basis of self-training. To obtain more accurate pseudo labels, UMT~\cite{UMT} transforms target domain data into source-like data in order to improve the quality of generated pseudo-labels. SimROD~\cite{Simrod} enhances the teacher model by augmenting its capacity for generating higher-quality pseudo boxes.

With the urgent need for data privacy protection, \emph{Source-Free Object Detection (SFOD)} has emerged as a new branch of UDAOD in recent years. Due to the complexity of the object detection task (numerous regions, multi-scale features, and complex network structure) and the challenge of the absent source data, simply applying the existing UDA-Classification or UDAOD methods to SFOD tasks can not get satisfied results~\cite{yuan2022simulation, li2022target}. Therefore, SFOD~\cite{SFOD} develops a novel framework that uses self-entropy descent to select high-quality pseudo labels for self-training. SOAP~\cite{SOAP} devises domain perturbation on the target data to help the model learn domain-invariant features that are invariant to the perturbations. LODS~\cite{LODS} proposes a style enhancement module and graph alignment constraint to help the model learn domain-independent features. A$^2$SFOD~\cite{AASFOD} divides target images into source-similar and source-dissimilar images and then adopts adversarial alignment between the teacher and student models. IRG~\cite{IRG} designs an instance relation graph network combined with contrastive loss to guide the contrastive representation learning.
While the majority of these approaches rely on the MT framework~\cite{run2023lin}, they tend to overlook the issue of training instability arising from a single teacher model. This oversight allows errors to be replicated by the student model, consequently constraining its performance. To tackle this concern, we propose a \emph{Periodically Exchange Teacher-Student} approach that leverages knowledge from historical models to prevent catastrophic forgetting for MT framework.

\begin{figure*}[t]
\begin{center}   \includegraphics[width=1.0\linewidth]{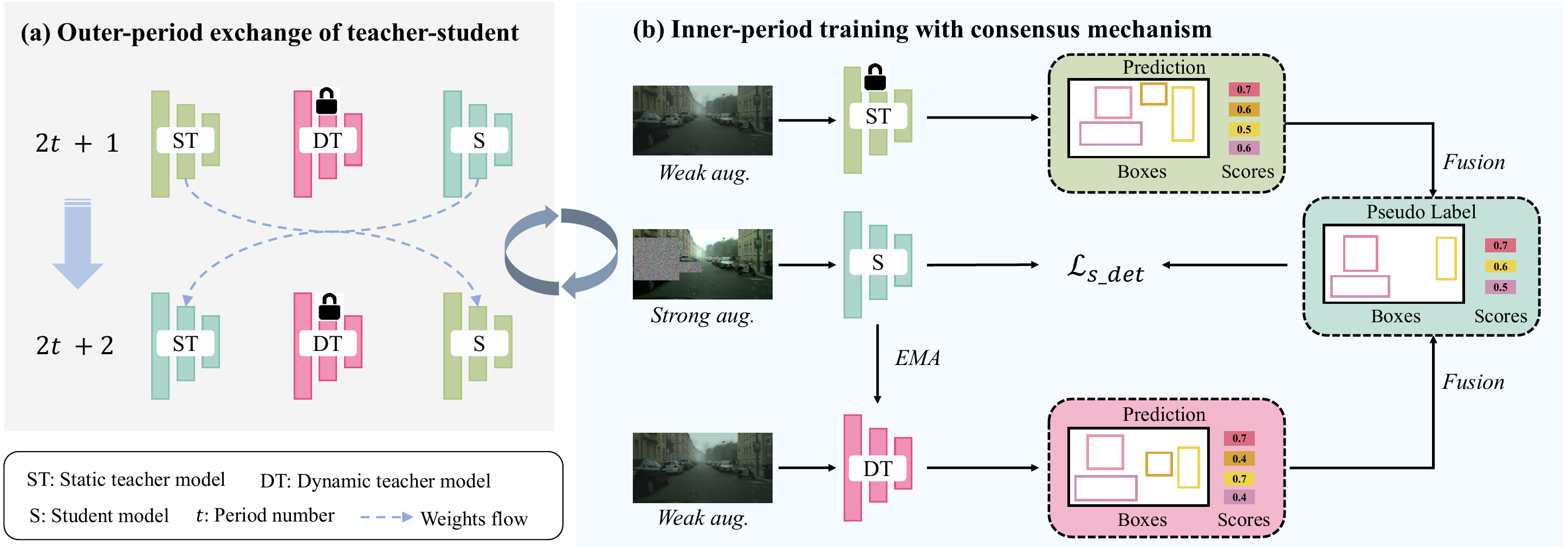}
\end{center}
    \vspace{-0.3cm}
   \caption{The training pipeline of the proposed \emph{Periodically Exchange Teacher-Student} method, which can be divided into two parts: (a) Outer-period exchange of teacher-student: exchange the weights between the student and static teacher after each period; (b) Inner-period training with consensus mechanism: update the dynamic teacher with an EMA of the student model, and train the student model with a consensus mechanism that fusions the predictions from multiple teachers.}
    \label{fig:2}
    \vspace{-0.3cm}
\end{figure*}

\section{Preliminary}
Let $\mathcal{D}_S=(\mathcal{X}_S,\mathcal{Y}_S)$ represent the labeled data in the source domain, and $\mathcal{D}_T=(\mathcal{X}_T)$ denote the unlabeled data in the target domain, where 
$\mathcal{X}_S=\{x_{s}^{i}\}^{N_{S}}_{i=1}$ represents the image set of the source domain, $\mathcal{Y}_S=\{y_{s}^{i}\}^{N_{S}}_{i=1}$ represents the corresponding label set containing object locations and category assignments for each image, and $\mathcal{X}_T=\{x_{t}^{i}\}^{N_{T}}_{i=1}$ denotes the image set of the unlabeled target domain. 
$N_s$ and $N_{t}$ correspond to the number of labeled source data and unlabeled target data, respectively.

In the setting of SFOD task, a source pre-trained model, denoted as $f_S:\mathcal{X}_S \rightarrow \mathcal{Y}_S$, is initially available to perform adaptation on unlabeled target domain. However, due to the inherent domain gap between the source and target domains, the mapping $f_S$ diminishes performance when directly applied to the target domain. Consequently, the primary objective of SFOD is to acquire a new mapping $f_T:\mathcal{X}_T \rightarrow \mathcal{Y}_T$ by leveraging the source-pretrained model $f_S$ in conjunction with the unlabeled target data $\mathcal{X}_T$ without accessing any source data.

Most previous SFOD methods use Faster-RCNN~\cite{faster_r-cnn} as their backbone network. To ensure a fair comparison with previous methods, we also adopt Faster-RCNN as the backbone network here. Therefore, the training goal of $f_T$ is similar to Faster-RCNN, which can be written as:
\begin{equation}
  \mathcal{L}_{det} = \mathcal{L}_{cls}^{RPN} + \mathcal{L}_{reg}^{RPN} + \mathcal{L}_{cls}^{ROI} + \mathcal{L}_{reg}^{ROI},
  \label{eq:det}
\end{equation} 
where $\mathcal{L}_{cls}^{RPN}$ and $\mathcal{L}_{reg}^{RPN}$ represent the losses of foreground prediction and box location from the RPN network, respectively. $\mathcal{L}_{cls}^{ROI}$ and $\mathcal{L}_{reg}^{ROI}$ are the losses of category prediction and box location from the ROI head, respectively.

\section{Methodology}
\subsection{Overview}
Our method involves using a static teacher model, a dynamic teacher model, and a student model. The pseudo code of training process can be seen in Algorithm \ref{alg:train}. Figure \ref{fig:2} shows the training pipeline of our method, which can be divided into two parts:
\begin{itemize}
\setlength{\itemsep}{1pt}
\item[1)] Outer-period exchange of teacher-student: After each period of training, we exchange the weights between the student model and the static teacher model. In other words, the static teacher and the student reverse their roles per period, as shown in Figure \ref{fig:2}(a). Note that the term ``period'' is synonymous with the concept of an epoch during training.
\item[2)] Inner-period training with consensus mechanism: The weights of static teacher model are fixed within each period. The dynamic teacher is updated by the EMA of the student model in each iteration, and the student model is supervised by the pseudo labels merged from the dynamic and static teacher models with consensus mechanism, as shown in Figure \ref{fig:2}(b).
\end{itemize}

\paragraph{Notations} For better understanding our method, we use $\Theta_{S}$, $\Theta_{ST}$ and $\Theta_{DT}$ to denote the student model, the static teacher model and the dynamic teacher model, respectively.

\begin{algorithm}[t]
\small
\caption{Python-like code of training process}
\label{alg:train}

\definecolor{codeblue}{rgb}{0.25,0.5,0.5}
\definecolor{codegreen}{rgb}{0,0.6,0}
\definecolor{codekw}{RGB}{207,33,46}
\lstset{
  backgroundcolor=\color{white},
  basicstyle=\fontsize{7.5pt}{7.5pt}\ttfamily\selectfont,
  columns=fullflexible,
  breaklines=true,
  captionpos=b,
  commentstyle=\fontsize{7.5pt}{7.5pt}\color{codegreen},
  keywordstyle=\fontsize{7.5pt}{7.5pt}\color{codekw},
  escapechar={|}, 
}

\begin{lstlisting}[language=python]
# Outer-period exchange of teacher-student
if epoch % time_period == 0:
    exchange_weight(student, static_teacher)

# Inner-period training with consensus mechanism
for _, images in enumerate(loader):
    # images: [N, C, H, W]
    # N: number of images per mini-batch
    # pre-process images by data augmentation
    img_w = weak_aug(images)
    img_s = strong_aug(img_w)
    
    # obtain predictions  
    pred_s = student(img_s)
    pred_st = static_teacher(img_w)     
    pred_dt = dynamic_teacher(img_w) 
    
    # produce pseudo label
    pseudo_labels = consensus(pred_st, pred_dt)
    
    # compute detection loss 
    loss = compute_loss(pred_s, pseudo_labels)
    
    # update the student by back-propagation
    loss.backward()
    
    # update the dynamic teacher by EMA
    update_teacher(student, dynamic_teacher)
\end{lstlisting}
\end{algorithm}

\subsection{Outer-period Exchange of Teacher-Student}
The training process can be divided into multiple independent time periods (i.e., epochs). Each period is represented as $t$.
At the $2t+2$ period, the weights of the student model are swapped by that of the static teacher model at the $2t+1$ period. Conversely, the weights of the static teacher model at the $2t+2$ period are exchanged by that of the student model at previous period. The exchange process can be written as:
\begin{align} 
\Theta_{S}^{2t+1}
 \longrightarrow 
 \Theta_{ST}^{2t+2}, \quad
\Theta_{ST}^{2t+1} \longrightarrow \Theta_{S}^{2t+2}, 
\end{align}
where $\Theta_{ST}^{2t+2}$ and $\Theta_{S}^{2t+2}$ denote the static teacher model and the student model at the $2t+2$ period, respectively. This exchange strategy keeps periodically recycling during the whole training process. 
 
The exchange strategy benefits each model from the following perspectives: 1) Student model: The static teacher model serves as a performance lower bound for the student model. If the student model crashes into a collapse issue guided by the declined dynamic teacher, the exchange can ensure that the student model reverts to previous period, effectively mitigating its downward trend. In essence, the exchange helps prevent a rapid decrease in the performance lower bound of the student model, thus improving its robustness.
2) Static teacher model: The exchange strategy ensures periodic updating to the static teacher model's knowledge, which is executed at a notably slow rate to enable a more stable model.
3) Dynamic teacher model: The dynamic teacher model is a temporal ensemble of the student model exchanged by the past student model. In practice, the updating rate of the dynamic teacher model is implicitly reduced.
Thus, it has a better ability to resist noise compared to the conventional mean-teacher framework~\cite{mean_teacher}. 
In summary, our \emph{periodically exchange teacher-student} strategy can enable the student and teacher models to mutually prevent catastrophic forgetting and uncontrollable collapse, thus improving the detection performance.

\subsection{Inner-period Training with Consensus Mechanism}
During each period, the static teacher maintains fixed weights until iterating to next period. Simultaneously, the dynamic teacher model is updated by the temporal ensembling of the student model, and the student model is updated by pseudo labels as supervision signals, where the pseudo labels are generated by combining the predictions of the dynamic and static teachers through the consensus mechanism. This procedure is illustrated in Figure \ref{fig:2}(b). The following sections delve into the details of the consensus mechanism, the learning process of the student model, and the updating of the dynamic teacher model.

\subsubsection{Consensus Mechanism}
\label{sec:consensus}
Our framework incorporates two distinct teachers: the static teacher and the dynamic teacher. A notable advantage of our approach is the ability to leverage predictions from both teachers to enhance the quality of pseudo labels. To this end, we design a consensus mechanism that includes two main steps: filtering and fusion. 
\vspace{-0.3cm}
\paragraph{Filtering} Since the output of teacher models contains inevitable noise (low-confidence predictions), we set a category confidence threshold $\delta=0.5$ to pre-filter low-confidence predictions. This can prevent the subsequent fusion process suffering from the interference of noisy labels.
\vspace{-0.6cm}
\paragraph{Fusion}
For a weakly-augmented target image $x_t \in \mathcal{X}_T$, the predictions of the static teacher and dynamic teacher are represented as $Y_{ST}=\{(b_{ST}^{i},c_{ST}^{i}, y_{ST}^{i})\}^{n}_{i=0}$ and $Y_{DT}=\{(b_{DT}^{j},c_{DT}^{j}, y_{DT}^{j})\}^{m}_{j=0}$, where $b, c, y$ represent the bounding box coordinates, classification confidence and category label of each predicted object, and $n, m$ denote the number of predicted objects of the static teacher and the dynamic teacher, respectively. Then, we select the objects with identical category and a higher intersection over union (IOU) between the predicted boxes of the static teacher and the dynamic teacher. The selection criterion can be represented as $IOU(b_{ST}^{i}, b_{DT}^{j}) \geq \eta \And y_{ST}^{i} = y_{DT}^{j}$,  where $\eta$ is the threshold of judging whether the predicted box belongs to the same object. We usually set $\eta=0.5$. Lastly, we employ the weighted boxes fusion (WBF) strategy~\cite{wbf} to merge the selected boxes derived from both the static teacher and dynamic teacher models. The process can be formulated as:
\begin{equation}
\begin{aligned}
&\widetilde{b} = \frac{1}{C}(\sum_{i=1}^{N} c_{ST}^{i} * b_{ST}^{i} + \sum_{j=1}^{M} c_{DT}^{j} *b_{DT}^{j}),
\\
&\widetilde{c} = \frac{\beta}{N} \sum_{i=1}^{N} c_{ST}^{i} + \frac{1 - \beta}{M}\sum_{j=1}^{M} c_{DT}^{j},
\end{aligned}
\label{eq:wbf}
\end{equation}
where $N, M$ are the number of boxes belonging to the same object predicted by the static teacher and the dynamic teacher, respectively. $C$ is the sum of $\sum_{i=1}^{N} c_{ST}^{i}$ and $\sum_{j=1}^{M} c_{DT}^{j}$. $\beta$ controls the fusion magnitude between the static teacher and dynamic teacher, which is ranged in $[0,1]$ and set to $0.5$ in this paper. We ultimately obtain pseudo label $\widetilde{Y}=\{( \widetilde{b},\widetilde{c},\widetilde{y})\}$ for the unlabeled target image $x_t$, where $\widetilde{b}$ and $\widetilde{c}$ denote the coordinates and confidence of the fused bounding box, respectively, and $\widetilde{y}$ is equivalent to $y_{ST}^{i}$.
The fused pseudo labels exhibit greater resistance to confirmation bias compared to those single-teacher framework.
\subsubsection{Student Learning}
Given an unlabeled target image $x_t$, its pseudo label can be represented as $\widetilde{Y}=\{(\widetilde{b}, \widetilde{y})\}$ that can be used as the supervision signal of the student model. Following Equation \ref{eq:det}, the training loss of the student model $\Theta_S$ can be defined as:
\begin{equation}
\begin{aligned}
  \mathcal{L}_{s\_{det}} = \sum\limits_{\bar{x}_t \in \mathcal{X}_T}
  &\mathcal{L}_{cls}^{RPN}(\Theta_S(\bar{x}_t), \widetilde{y}) + \mathcal{L}_{reg}^{RPN}(\Theta_S(\bar{x}_t), \widetilde{b}) + \\
  & \mathcal{L}_{cls}^{ROI}(\Theta_S(\bar{x}_t), \widetilde{y}) + \mathcal{L}_{reg}^{ROI}(\Theta_S(\bar{x}_t), \widetilde{b}),
\label{eq:student_learning}
\end{aligned}
\end{equation} 
where $\bar{x}_t$ denotes the strongly-augmented version of the target image $x_t$. Since the proposed consensus mechanism can provide more precise bounding boxes compared with previous studies~\cite{UMT}, we use both the category prediction loss and box location loss to train the student model. 
\subsubsection{Dynamic Teacher Updating}
\label{sec:teacher}
Throughout each period, the static teacher model maintains fixed weights across various iterations, whereas the dynamic teacher model adjusts its weights in each iteration. We follow the conventional MT framework that uses the exponential moving average (EMA) strategy to update the dynamic teacher model $\Theta_{DT}$. This can be formulated as:
\begin{equation}
  \Theta_{DT} \gets \alpha\Theta_{DT}' + (1 - \alpha)\Theta_{S},
  \label{eq:ema1}
\end{equation}
where $\Theta_{DT}$ represents the dynamic teacher in current iteration, while $\Theta_{DT}'$ pertains to the dynamic teacher in previous iteration. The hyper-parameter $\alpha$ controls the update rate of the dynamic teacher, with a higher value leading to a slower update rate. In this study, we empirically set $\alpha$ to 0.999.

\section{Experiments}
We conduct comprehensive experiments to evaluate the effectiveness of our method on multiple standard SFOD benchmarks. Then, we perform ablation studies by using different exchange strategies to stress the effectiveness of the proposed periodic exchange strategy. Finally, we analyze the promising results of our method through detailed visualization and component analysis.
\begin{table*}[t]
\centering
\setlength\tabcolsep{9pt}\resizebox{0.95\textwidth}{!}{\begin{tabular}{ll|cccccccc|c}
\toprule[1.0pt]
    \multicolumn{2}{c|}{Methods}                                      & Person         & Rider          & Car            & Truck          & Bus            & Train          & Motor         & Bicycle        & mAP   \\
\midrule[0.5pt]
\multicolumn{1}{c|}{} & Source only (Single level)  & 23.4   & 23.8  & 29.7   & 8.1  & 12.9 & 5.0          & 18.3    & 24.5    & 18.2 \\
\multicolumn{1}{c|}{} & Source only (All levels)    & 35.1   & 39.4  & 47.0   & 10.7  & 32.5 & 10.1          & 30.0    & 36.9    & 30.7 \\

\midrule[0.5pt]
    \multicolumn{1}{c|}{\multirow{5}{*}{UDAOD}} & MAF ~\cite{MAF}           & 28.2   & 39.5  & 43.9 & 23.8  & 39.9 & 33.3          & 29.2  & 33.9    & 34.0   \\
    \multicolumn{1}{c|}{}  &  SW-Faster ~\cite{SW}     & 32.3   & 42.2  & 47.3 & 23.7  & 41.3 & 27.8          & 28.3  & 35.4    & 34.8 \\
    \multicolumn{1}{c|}{}  & iFAN  ~\cite{ifan}         & 32.6   & 40.0    & 48.5 & 27.9  & 45.5 & 31.7          & 22.8  & 33.0      & 35.3 \\
    \multicolumn{1}{c|}{}  & CR-DA-DET  ~\cite{CR-DA-DET}    & 32.9   & 43.8  & 49.2 & 27.2  & 45.1 & 36.4          & 30.3  & 34.6    & 37.4 \\
    \multicolumn{1}{c|}{}  & AT-Faster ~\cite{AT} & 34.6   & 47.0    & 50.0   & 23.7  & 43.3 & \textbf{38.7} & 33.4  & 38.8    & 38.7 \\
\midrule[0.5pt]
    \multicolumn{1}{c|}{\multirow{7}{*}{SFOD}}  & SED(Mosaic)~\cite{SFOD} & 33.2  & 40.7  & 44.5  & 25.5  & 39.0 & 22.2  & 28.4  & 34.1  & 33.5 \\
    \multicolumn{1}{c|}{}  & HCL~\cite{HCL}    & 26.9  & 46.0    & 41.3  & \textbf{33.0}    & 25.0    & 28.1  & 35.9  & 40.7  & 34.6 \\
    \multicolumn{1}{c|}{}   & $\text{A}^2$SFOD ~\cite{AASFOD} &  32.3 & 44.1 &  44.6 & 28.1 & 34.3 & 29.0 & 31.8 & 38.9 & 35.4 \\
    \multicolumn{1}{c|}{}  & SOAP ~\cite{SOAP} & 35.9   & 45.0    & 48.4 & 23.9  & 37.2 & 24.3          & 31.8  & 37.9    & 35.5 \\
    \multicolumn{1}{c|}{}  & LODS ~\cite{LODS} & 34.0     & 45.7  & 48.8 & 27.3  & 39.7 & 19.6          & 33.2  & 37.8    & 35.8 \\
    \multicolumn{1}{c|}{}  &  IRG~\cite{IRG}   & 37.4  & 45.2  & 51.9  & 24.4  & 39.6  & 25.2  & 31.5  & 41.6  & 37.1 \\
    \multicolumn{1}{c|}{}   & Ours (Single level) & 42.0 & 48.7 & 56.3 &  19.3 &  39.3 &  5.5 &  34.2 & 41.6 &  35.9 \\
    \multicolumn{1}{c|}{}   & Ours (All levels) & \textbf{46.1} & \textbf{52.8} & \textbf{63.4} &  21.8 &  \textbf{46.7} &  5.5 &  \textbf{37.4} &  \textbf{48.4} &  \textbf{40.3} \\
\midrule[0.5pt]
\multicolumn{1}{c|}{} & Oracle         & 51.3   & 57.5  & 70.2 & 30.9  & 60.5 & 26.9          & 40.0  & 50.4    & 48.5 \\
\bottomrule[1.0pt]
\end{tabular}
}
\vspace{-0.2cm}
\normalsize
\caption{Results of adaptation from normal to foggy weather (C2F). “Source only” and “Oracle” refer to the models trained by only using labeled source domain data and labeled target domain data, respectively.}
\label{tab:C2F}
\vspace{-0.2cm}
\end{table*}

\begin{table*}[t]
\centering
\setlength\tabcolsep{9pt}\resizebox{0.95\textwidth}{!}{
\begin{tabular}{ll|ccccccc|c}
\toprule[1.0pt]
 \multicolumn{2}{c|}{Methods}            & Truck & Car  & Rider & Person & Motor & Bicycle & Bus  & mAP  \\
\midrule[0.5pt]
\multicolumn{1}{c|}{} &  Source only        & 9.9   & 51.5 & 17.8  & 28.7   & 7.5   & 10.8    & 7.6  & 19.1 \\
\midrule[0.5pt]
\multicolumn{1}{c|}{\multirow{3}{*}{UDAOD}} & DA-Faster~\cite{DA-Faster}          & 14.3  & 44.6 & 26.5  & 29.4   & 15.8  & 20.6    & 16.8 & 24.0   \\
\multicolumn{1}{c|}{} & SW-Faster~\cite{SW}          & 15.2  & 45.7 & 29.5  & 30.2   & 17.1  & 21.2    & 18.4 & 25.3 \\
\multicolumn{1}{c|}{} & CR-DA-DET~\cite{CR-DA-DET} & 19.5  & 46.3 & 31.3  & 31.4   & 17.3  & 23.8    & 18.9 & 26.9 \\
\midrule[0.5pt]
\multicolumn{1}{c|}{\multirow{4}{*}{SFOD}} & SED~\cite{SFOD} & 20.4  & 48.8 & 32.4  & 31.0     & 15.0    & 24.3    & 21.3 & 27.6 \\
\multicolumn{1}{c|}{} & SED(Mosaic)~\cite{SFOD}       & {20.6}    & {50.4}    & 32.6          & 32.4          & {18.9}    & 25.0            & {23.4}    & 29.0            \\
\multicolumn{1}{c|}{} & $\text{A}^2$SFOD ~\cite{AASFOD}& \textbf{26.6} & 50.2          & \textbf{36.3} & {33.2}    & \textbf{22.5} & \textbf{28.2} & \textbf{24.4} & \textbf{31.6} \\
\multicolumn{1}{c|}{} & Ours             & 19.3          & \textbf{62.4} & {34.5}    & \textbf{42.6} & 17.0          & {26.3}    & 16.9          & {31.3}    \\
\midrule[0.5pt]
\multicolumn{1}{c|}{} & Oracle & 47.7  & 72.1  & 38.4  & 50.0    & 25.5  & 32.3  & 42.8 &44.1 \\
\bottomrule[1.0pt]
\end{tabular}
}
\vspace{-0.2cm}
\caption{Results of adaptation from small-scale to large-scale dataset (C2B).}
\vspace{-0.5cm}
\label{tab:C2B}
\end{table*}
\begin{table}[t]
  \centering
\setlength\tabcolsep{9pt}\resizebox{0.48\textwidth}{!}{
 \begin{tabular}{l|c|l|c}
    \toprule
    Methods & mAP & Methods & mAP \\
    \midrule[0.5pt]
    Source only & 36.3  & MeGA-CDA~\cite{MeGA} & 43.0 \\
    DA-Faster~\cite{DA-Faster} & 38.5  & NL~\cite{NL}    & 43.0 \\
    SW-Faster~\cite{SW} & 37.9  & SAPNet~\cite{SAPNet} & 43.4 \\
    MAF~\cite{MAF}   & 41.0    & SGA-S~\cite{SGA-S} & 43.5 \\
    AT-Faster~\cite{AT} & 42.1  & CST-DA~\cite{CST-DA} & 43.6 \\
    \midrule[0.5pt]
    SOAP~\cite{SOAP} & 42.7  & $\text{A}^{2}$SFOD~\cite{AASFOD} & 44.9 \\
    SFOD~\cite{SFOD}  & 43.6  & IRG~\cite{IRG}   & 45.7 \\
    LODS~\cite{LODS}  & 43.9  & Ours  & \textbf{47.0} \\
    SED(Mosaic)~\cite{SFOD} & 44.6  & Oracle & 68.9 \\
    \bottomrule[1.0pt]
    \end{tabular}%
    }
    \vspace{-0.2cm}
    \caption{Results of adaptation across cameras (K2C).}
    \vspace{-0.1cm}
  \label{tab:K2C}%
\end{table}%

\begin{table}[t]
  \centering
\setlength\tabcolsep{9pt}\resizebox{0.48\textwidth}{!}{
    \begin{tabular}{l|c|l|c}
    \toprule[1.0pt]
    \multicolumn{1}{c|}{Methods} & mAP   & \multicolumn{1}{c|}{Methods} & mAP \\
    \midrule[0.5pt]
     Source only & 40.5  & NL~\cite{NL}    & 43.0 \\
    MAF~\cite{MAF}   & 41.1  & UMT~\cite{UMT}   & 43.1 \\
    AT-Faster~\cite{AT}& 42.1 & MeGA-CDA~\cite{MeGA}  & 44.8 \\
     HTCN~\cite{HTCN}  & 42.5  & CR-DA-DET~\cite{CR-DA-DET} & 46.1 \\
    \midrule
    SED~\cite{SFOD} & 42.3  & $\text{A}^2$SFOD~\cite{AASFOD} & 44.0 \\
    SED(Mosaic)~\cite{SFOD} & 43.1  & Ours  & \textbf{57.8} \\
    IRG~\cite{IRG} & 43.2  & Oracle & 68.9 \\
    \bottomrule[1.0pt]
    \end{tabular}%
    }
    \vspace{-0.2cm}
      \caption{Results of adaptation from synthetic to real scenes (S2C).}
    \vspace{-0.1cm}
  \label{tab:s2c}%
\end{table}%

\subsection{Experimental Setup}
\paragraph{Task Settings.} Following the existing works~\cite{SFOD,AASFOD}, we validate our method on the four popular SFOD tasks which represent different types of domain shift, including 1) Cityscapes-to-Foggy-Cityscapes (C2F): Adaptation from normal to foggy weather. 2) Cityscapes-to-BDD100k (C2B): Adaptation from small to large-scale dataset. 3) KITTI-to-Cityscapes-Car (K2C): Adaptation across different cameras. 4) Sim10k-to-Cityscapes-Car (S2C): Adaptation from synthetic to real images. The A-to-B represents the adaption of the model pre-trained on the source domain A to the target domain B.

\paragraph{Datasets.}
There are five datasets used in the aforementioned tasks:
1) \emph{Cityscapes}~\cite{cityscapes} is a street view dataset containing 5,000 images with instance-level pixel annotation from different cities in different seasons, where 2,925 training images and 500 validation images are used in the following experiments.
2) \emph{Foggy Cityscapes}~\cite{cityscapes_foggy} is also a street view dataset similar to Cityscapes, but its images are processed by three levels (0.005, 0.01, 0.02) of artificial simulation of extreme foggy scenes.
3) \emph{KITTI}~\cite{kitti} is a widely used benchmark dataset for autonomous driving which contains many images from different real-world street scenes. There are only 7,481 training images used in the experiments.
4) \emph{SIM10k}~\cite{sim10k} is a synthetic dataset consisting of  10,000 city scenery images of cars.
5) \emph{BDD100k}~\cite{bdd100k} is a large-scale open source video dataset for autonomous driving, including 100k images from different times, different weather conditions and driving scenarios.

\subsection{Implementation Details}
    \label{exp:details}
    Our method is implemented based on PyTorch platform using detectron2 framework~\cite{detectron2}. Following the previous study~\cite{SFOD}, we use Faster-RCNN~\cite{faster_r-cnn} with the backbone of VGG16 pre-trained on the ImageNet as the base detection model in our method.
    All images are scaled by resizing the shorter edge of the image to 600 pixels before training. The data augmentation strategy includes random erasing, random horizontal flip, and color transformation.
    We adopt the SGD as the optimizer with an initial learning rate of 8e-4, a decay rate of 0.1. The batch size is set to 8.

    The training process of our method consists of two stages: warm-up and adaptation. In the warm-up stage, the learning rate increases gradually from 0 to 8e-4. The static teacher model freezes its weights and the dynamic teacher model keeps updating during the first two epochs. In the fine-tuning stage, the weights of the student model and the static teacher model are exchanged per epoch, and the EMA rate of the dynamic teacher model is set to 0.999. During evaluation process, we reserve the dynamic teacher model for inference and choose the mean average precision (mAP) with an IOU threshold of 0.5 as the evaluation measure.

\begin{table}[t]
  \centering
\setlength\tabcolsep{9pt}\resizebox{0.48\textwidth}{!}{
    \begin{tabular}{c|c|c|c|c}
    \toprule[1.0pt]
     Foggy level     & Method & DT    & ST   & mAP \\
    \midrule[0.5pt]
    \multirow{4}[2]{*}{All levels} & Source only &  -    &  -     & 30.7 \\
  &    Single-teacher       &   -    & \checkmark   & 36.6 \\
       &     Single-teacher      & \checkmark &   -       & 38.0  \\
          & Ours  & \checkmark & \checkmark   &  \textbf{40.3} \\
    \midrule[0.5pt]
         \multirow{4}[1]{*}{Single level}     & Source only &  -    &  -     & 18.2 \\
&   Single-teacher    &   -    & \checkmark  & 27.2  \\
          &  Single-teacher      &  \checkmark &  -     &  32.9   \\
          & Ours   & \checkmark & \checkmark &   \textbf{35.9}  \\
    \bottomrule[1.0pt]
    \end{tabular}%
}  \vspace{-0.2cm}
    \caption{Results of single-teacher and multi-teacher methods on C2F benchmark. \textbf{DT} and \textbf{ST} represent the dynamic teacher and static teacher, respectively.}
    \label{tab:ablation2}%
\end{table}%

\begin{table}[t]
\centering
\setlength\tabcolsep{9pt}\resizebox{0.48\textwidth}{!}{
\begin{tabular}{c|c|c|c}
\toprule[1.0pt]
Weights flowing strategy & K2C & C2F & Avg\\
\midrule[0.5pt]
Baseline           & 43.8  & 36.6 & 40.2\\
S $\longrightarrow$ ST            & 46.8 & 39.6& 43.2\\
DT $\longrightarrow$  S            & 44.1 & 37.8& 41.0\\
DT $\longrightarrow$  ST          & 46.4 & 38.9& 42.7 \\
S $\longleftrightarrow$  ST (Ours)        & \textbf{47.0} & \textbf{40.3} & \textbf{43.7}\\
\bottomrule[1.0pt]
\end{tabular}
}
\vspace{-0.2cm}
\caption{Results of different exchange strategies on K2C and C2F benchmarks. ``Baseline'' means training the proposed multi-teacher framework without any weights flowing. }
\label{tab:abl1}
\end{table}

\begin{figure*}[ht]
\subfigure
{
    \begin{minipage}[b]{.23\linewidth}
        \centering
        \includegraphics[scale=.26]{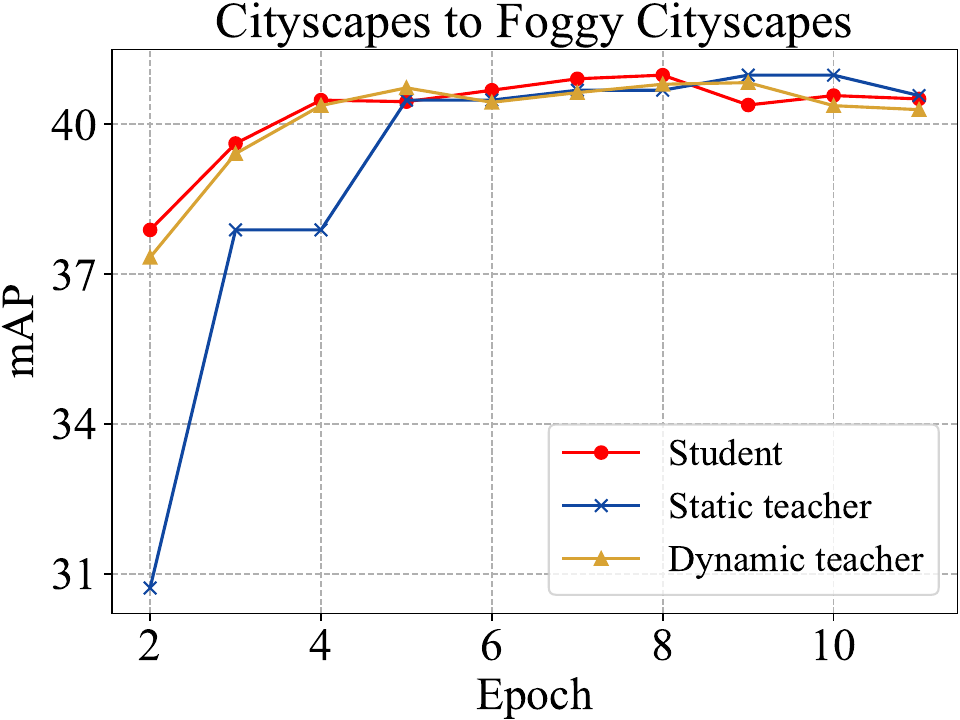}
    \end{minipage}
}
\subfigure
{
    \begin{minipage}[b]{.23\linewidth}
    \centering
        \includegraphics[scale=.26]{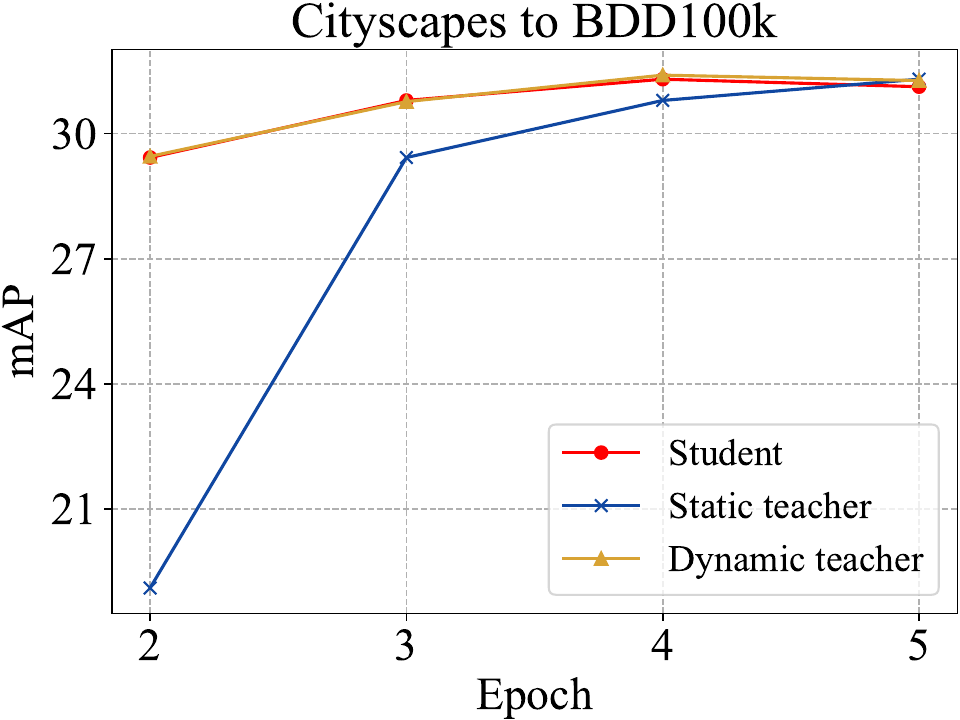} 
    \end{minipage}
}
\subfigure
{
    \begin{minipage}[b]{.23\linewidth}
    \centering
        \includegraphics[scale=.26]{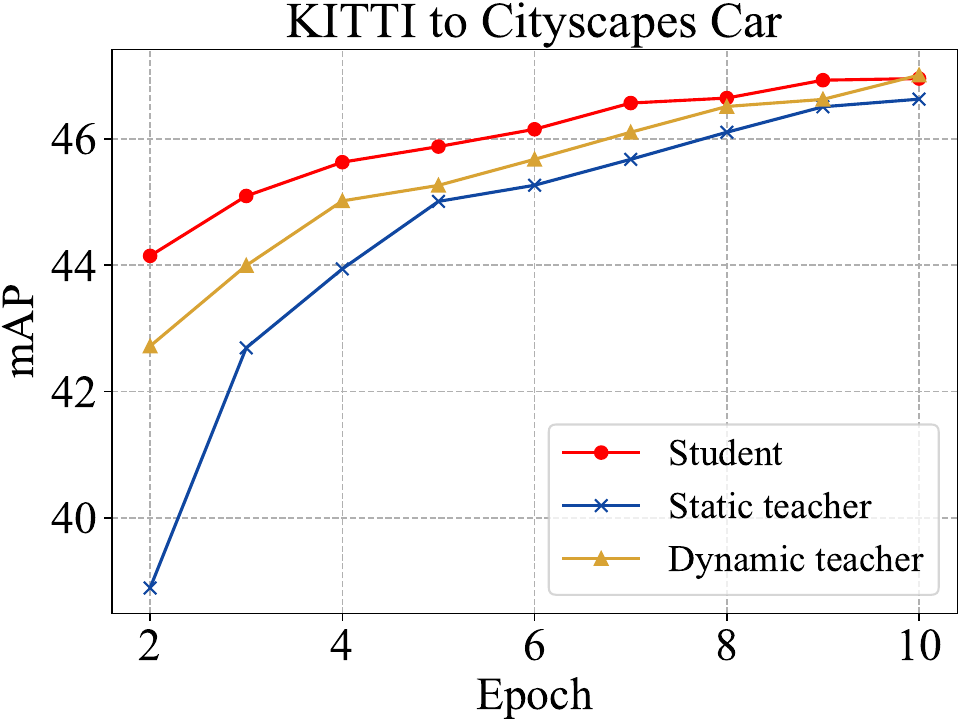}
    \end{minipage}
}
\subfigure
{
    \begin{minipage}[b]{.23\linewidth}
    \centering
        \includegraphics[scale=.26]{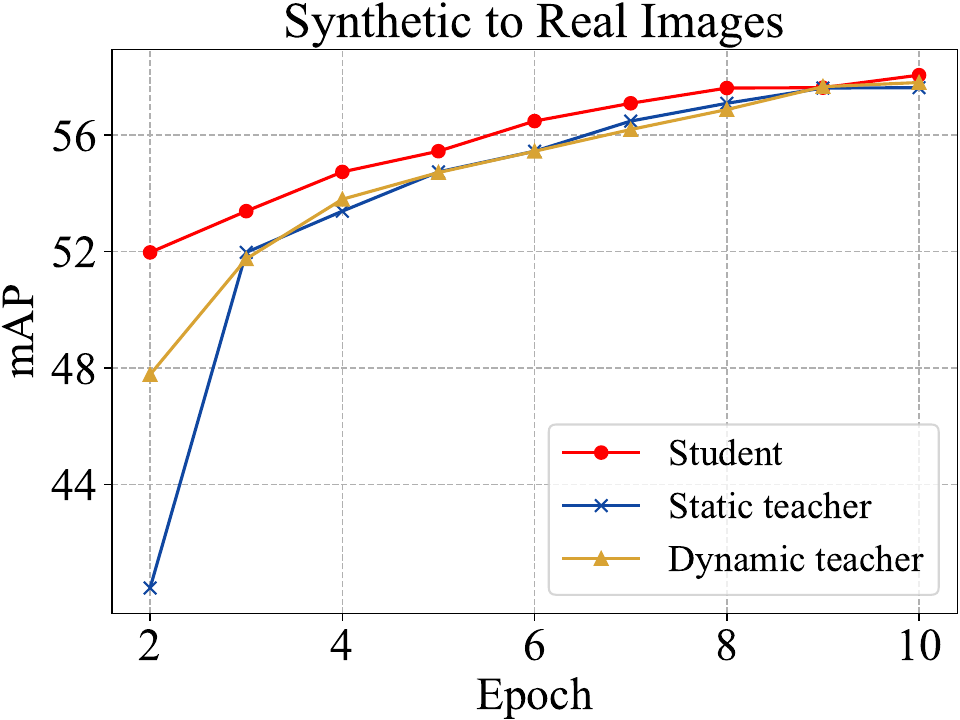}
    \end{minipage}
}
\vspace{-0.4cm}
\caption{The training curves of each model within the multi-teacher framework during the whole training process.}
\vspace{-0.4cm}
\label{fig:plot}
\end{figure*}

\subsection{Comparison with Existing SOTA Methods}
UDAOD and SFOD have a similar task setting. Therefore, we compare our method with existing UDAOD and SFOD methods. Table \ref{tab:C2F}-\ref{tab:s2c} show the comparison results, where ``Source only'' and ``Oracle'' represent the models which are only trained in source domain or target domain data, respectively. They represent the upper and lower performance bounds of the SFOD task.

\paragraph{C2F: Adaptation from Normal to Foggy Weather.}
In real-world application scenarios, e.g., automated driving, object detectors tend to encounter various complex weather conditions. To study the domain shift caused by weather conditions, we perform the adaptation from normal weather to foggy weather. For fair comparison, our experiments are conducted in two manners: 1) All levels: Using all target data with three foggy levels for training. 2) Single level: Using partial target data with a foggy level at 0.02 for training.  The results are shown in Table \ref{tab:C2F}. Our method achieves an mAP score of 40.3\%, which outperforms both the UDAOD and SFOD methods on this benchmark.

\paragraph{C2B: Adaptation from Small-scale to Large-scale Dataset.}
Annotating a large number of data for detection task can be very expensive and time-consuming. Therefore, the most economical way is to transfer knowledge from small-scale labeled datasets to large-scale unlabeled datasets. However, different datasets exhibit varying degrees of domain shifts. To validate the effectiveness of our method on such task, we transfer the source-pretrained model from Cityscapes (source domain) to BDD100k (target domain). Following the setting of previous studies~\cite{SFOD,AASFOD}, we keep 8 categories in BDD100k that are the same as Cityscapes. Since the detection performance of the category “train” is always close to 0, we only report the mAP score of 7 categories in Table \ref{tab:C2B}. The results show that our method achieves very competitive performance with the latest state-of-the-art SFOD method on this benchmark.
\vspace{-0.3cm}
\paragraph{K2C: Adaptation across Various Cameras.}
Due to different camera settings (e.g., angle, resolution, quality, and type), domain shifts always occur in cross-camera images.
To explore our method on cross-camera images, we adapt the model trained on KITTI to SIM10k, a dataset with images taken from real-world but different photographic equipment. Following previous studies, we only evaluate the performance on ``Car'' category. The results are reported in Table \ref{tab:K2C}, where we can see that our method obtains state-of-the-art performance on this benchmark.
\vspace{-0.3cm}
\paragraph{S2C: Adaptation from Synthetic to Real Scenarios.}
Synthetic images provide an alternative to address the challenges of data collection and manual labeling. However, there is a substantial domain gap between synthetic data and real data. To study the adaptation from synthetic to real scenes, we use the model pre-trained on the entire Sim10k dataset as the source model. The training set of Cityscapes is used as target data by reserving car images and discarding other categories. Results in Table \ref{tab:s2c} show that our method outperforms the existing SFOD approach by a large margin of +13.8\%, which demonstrates the superiority of our method on this benchmark.

\subsection{Ablation study}
\paragraph{Single-teacher VS. Multi-teacher.} 
We investigate the necessity of multi-teacher framework by comparing it with single-teacher method on C2F benchmark. The single-teacher methods employ either a static teacher or a dynamic teacher to guide the student learning process, which no longer involves using the exchange strategy and consensus mechanism. 
As shown in Table \ref{tab:ablation2}, our multi-teacher framework achieves the best performance compared to the single-teacher frameworks on both foggy levels. The success can be attributed to the superiority of exchange strategy and consensus mechanism in multi-teacher framework.

\paragraph{Weights Flowing Strategy.} 
To verify the effectiveness of the proposed method, we also explore the performance of other weights flowing strategies. The comparison results are shown in Table \ref{tab:abl1}, where $A \rightarrow B$ represents the single-direction weights flowing strategy that model B copies the weights of model A, while model A retains its weights, and $A \leftrightarrow B$ denotes our double-direction weights flowing strategy. We can see that all weights flowing strategies show the superiority to the baseline model that does not involve any weights swapping. Moreover, the proposed double-direction weights flowing strategy outperforms other single-direction strategies on both K2C and C2F benchmarks. This again demonstrates the superiority of our method.

\begin{figure*}[t]
\subfigure
{
    \rotatebox{90}{\scriptsize{~~~~~~~Source Only}}
    \begin{minipage}[b]{.235\linewidth}
        \centering
        \includegraphics[scale=.078]{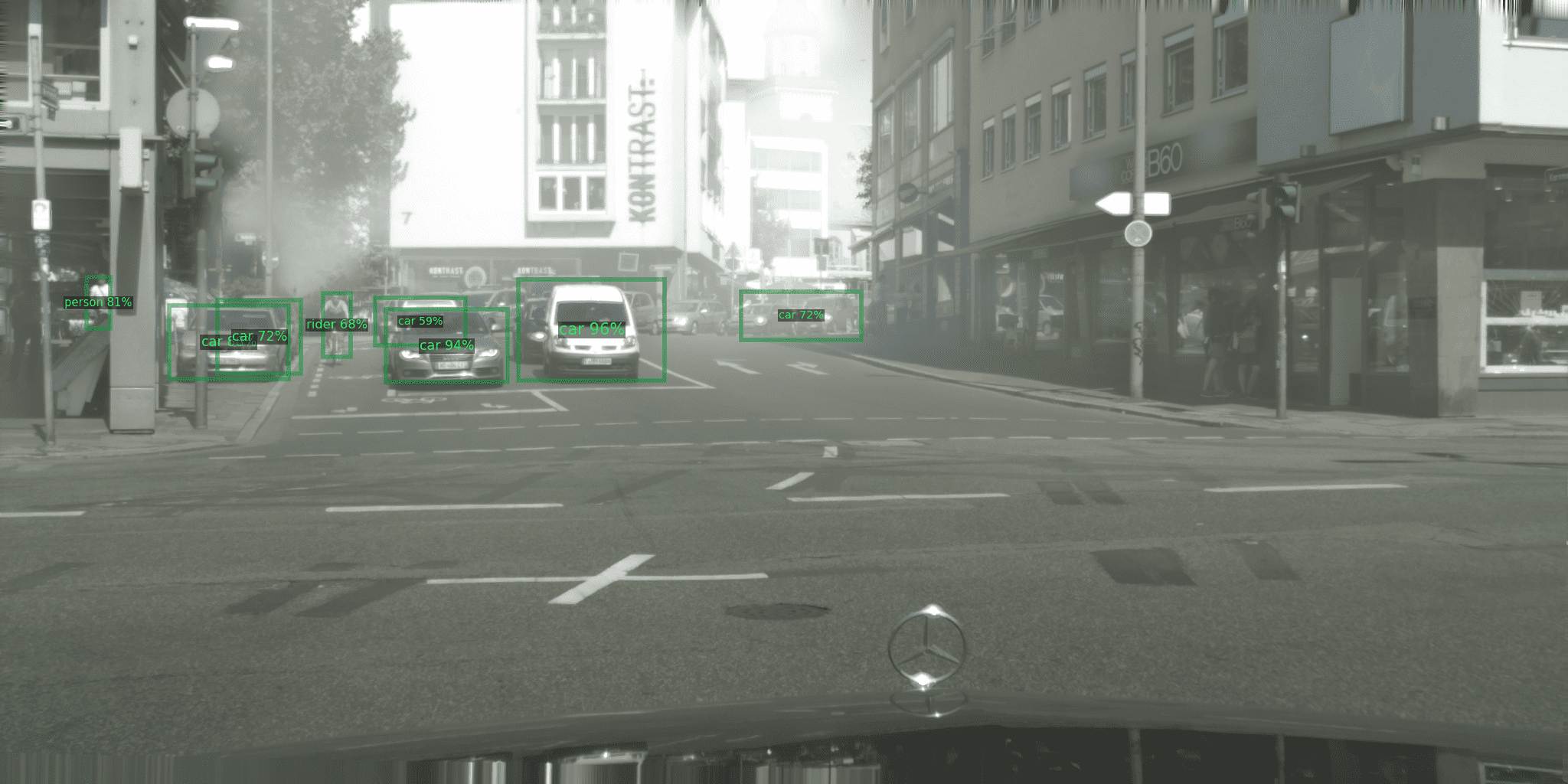}\vspace{2pt}
    \end{minipage}
    \begin{minipage}[b]{.235\linewidth}
        \centering
        \includegraphics[scale=.078]{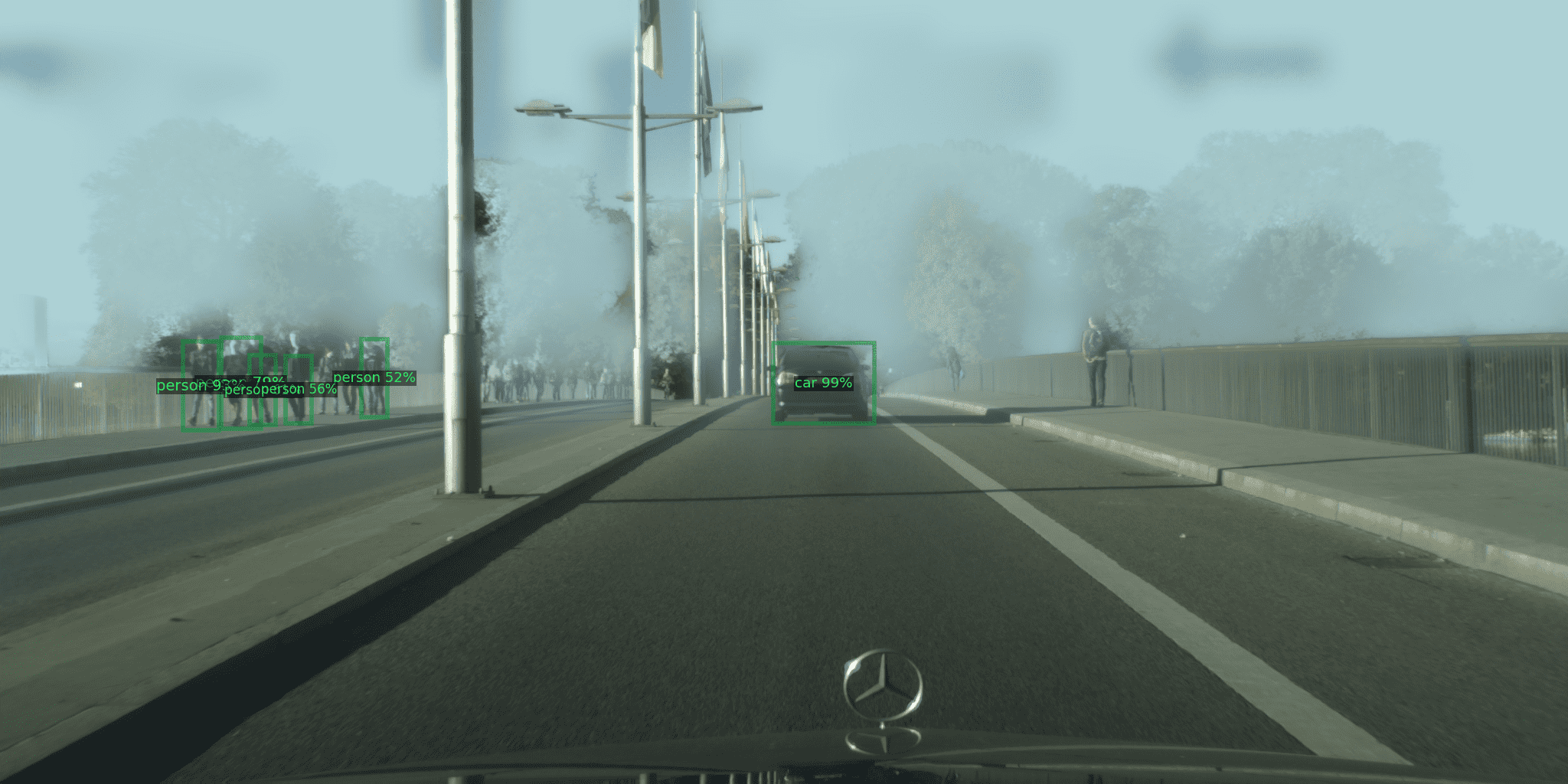} \\\vspace{2pt}
    \end{minipage}
    \begin{minipage}[b]{.235\linewidth}
        \centering
        \includegraphics[scale=.078]{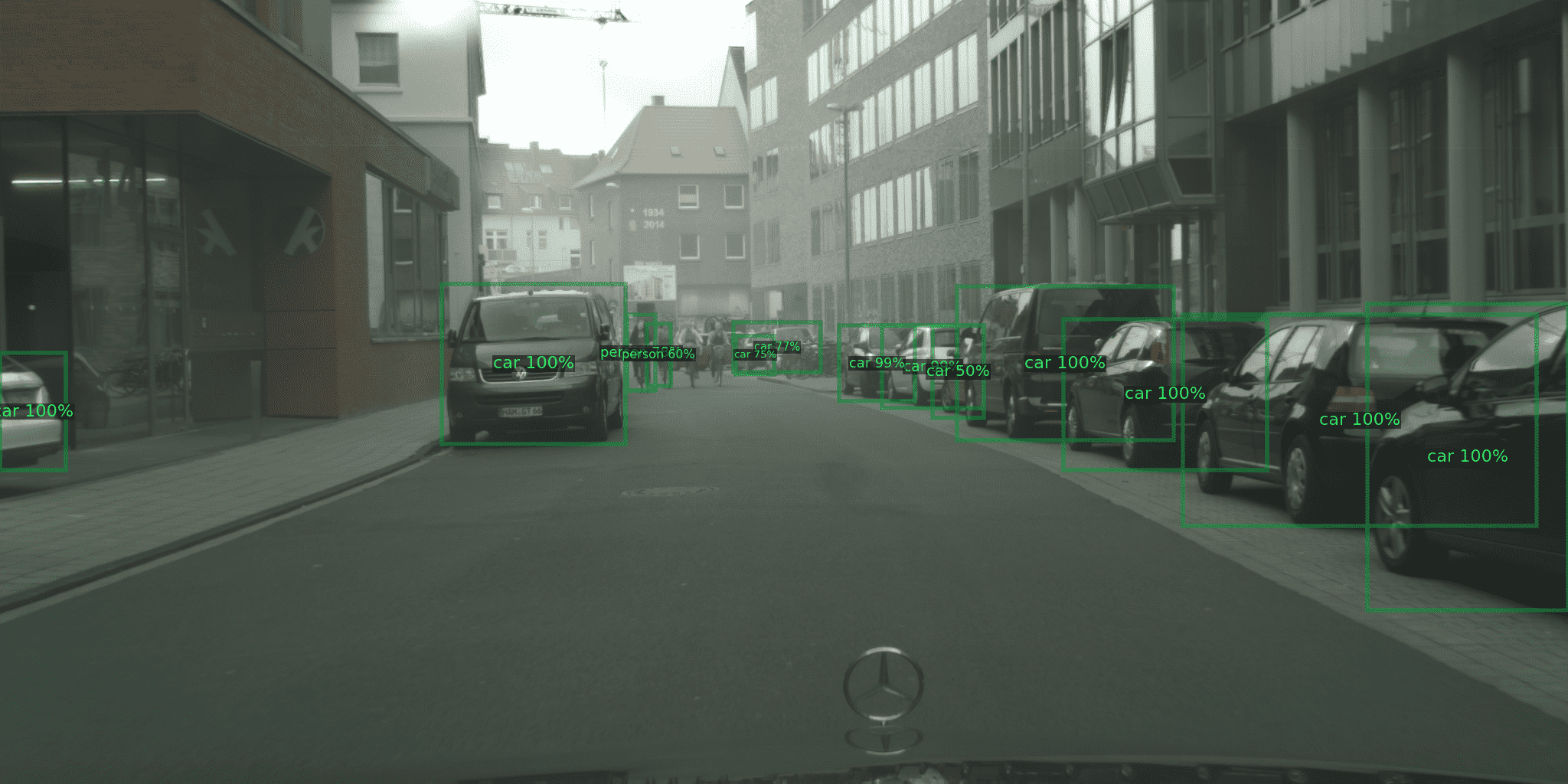} \\\vspace{2pt}
    \end{minipage}
    \begin{minipage}[b]{.235\linewidth}
        \centering  
        \includegraphics[scale=.078]{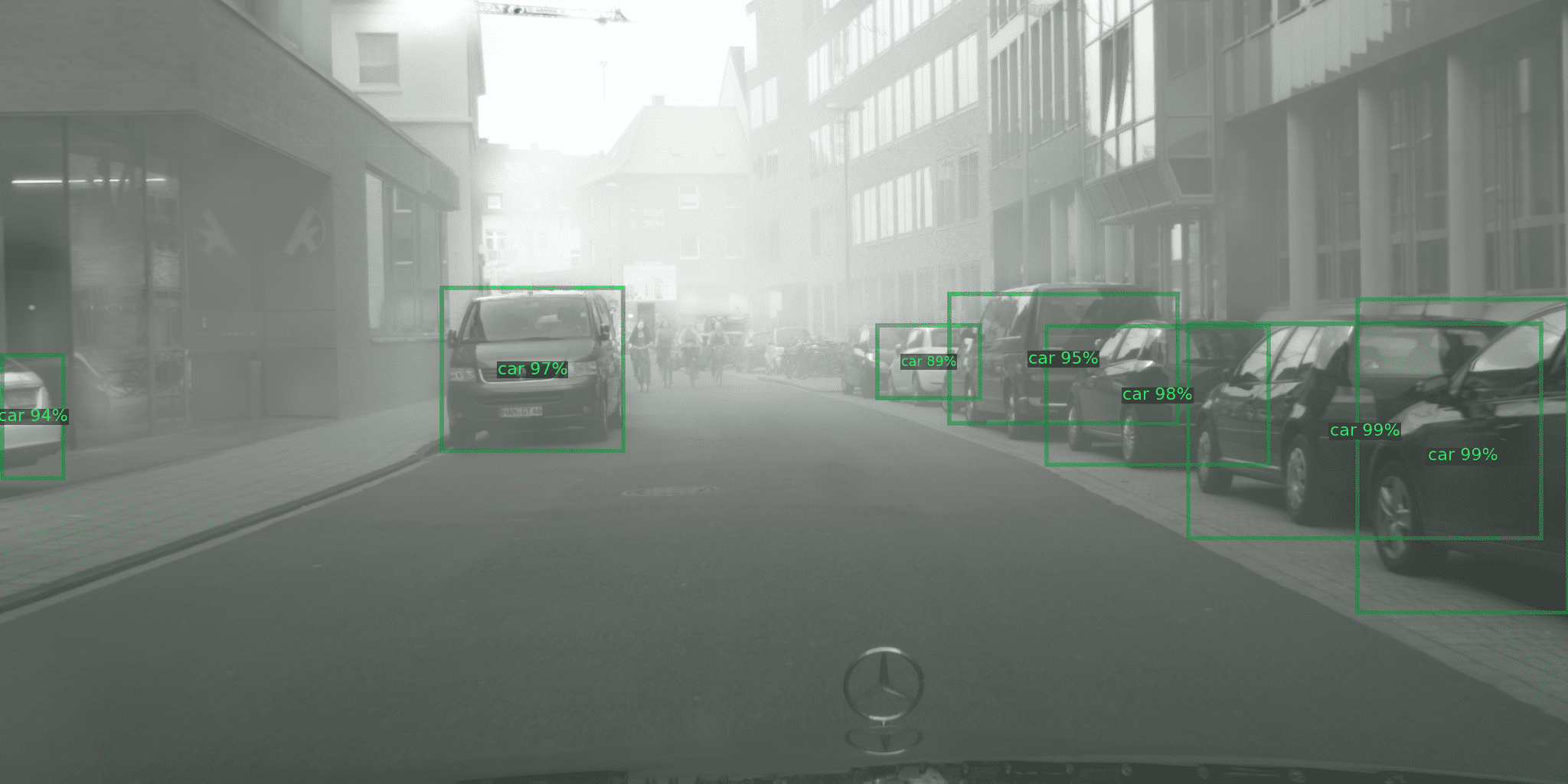} \\\vspace{2pt}
    \end{minipage}         
}
\\
\subfigure
{
    \rotatebox{90}{\scriptsize{~~~~~~~~~~ST (4999)}}
    \begin{minipage}[b]{.235\linewidth}
        \centering
        \includegraphics[scale=.078]{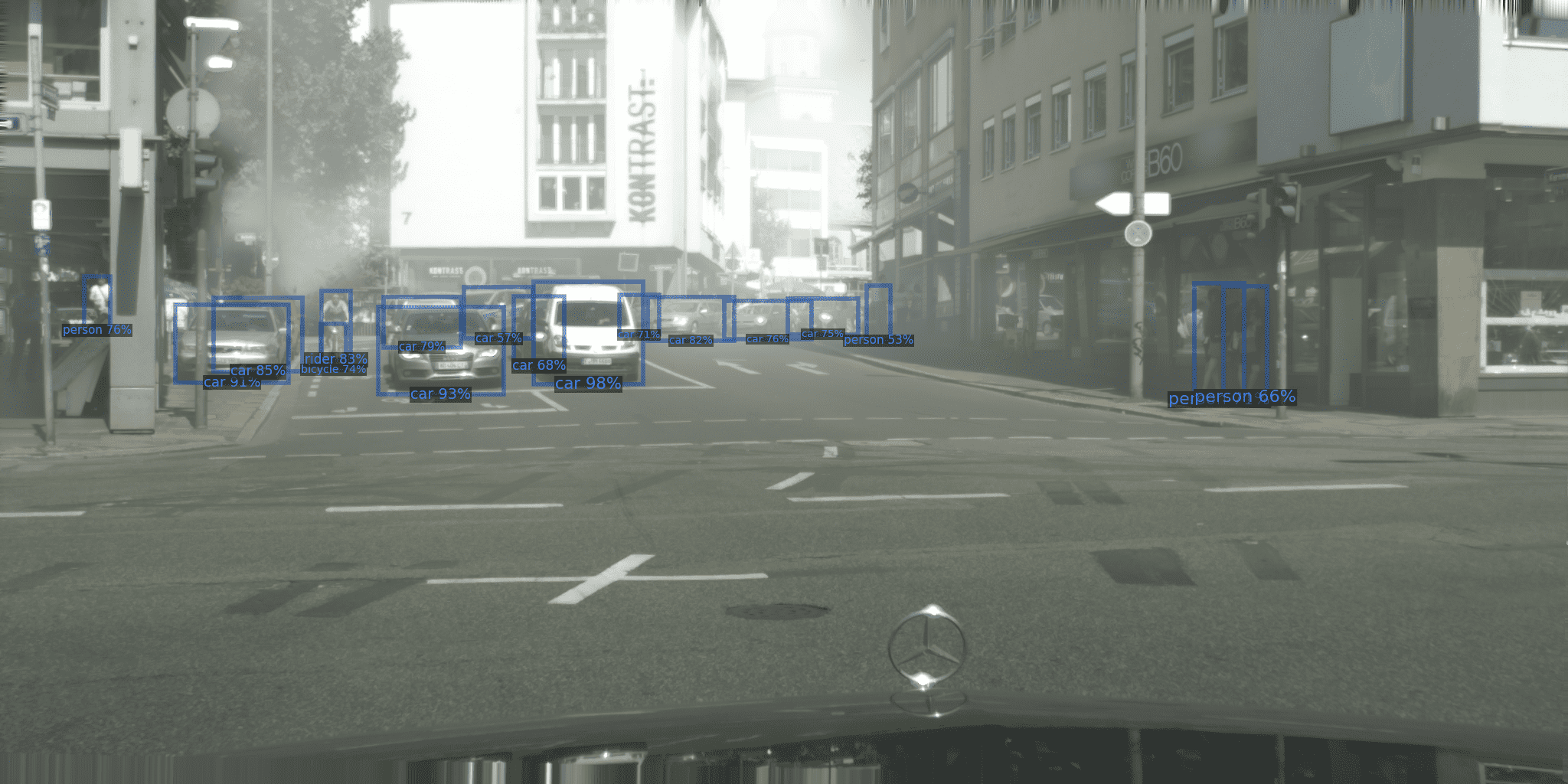}\vspace{2pt}
    \end{minipage}
    \begin{minipage}[b]{.235\linewidth}
        \centering
        \includegraphics[scale=.078]{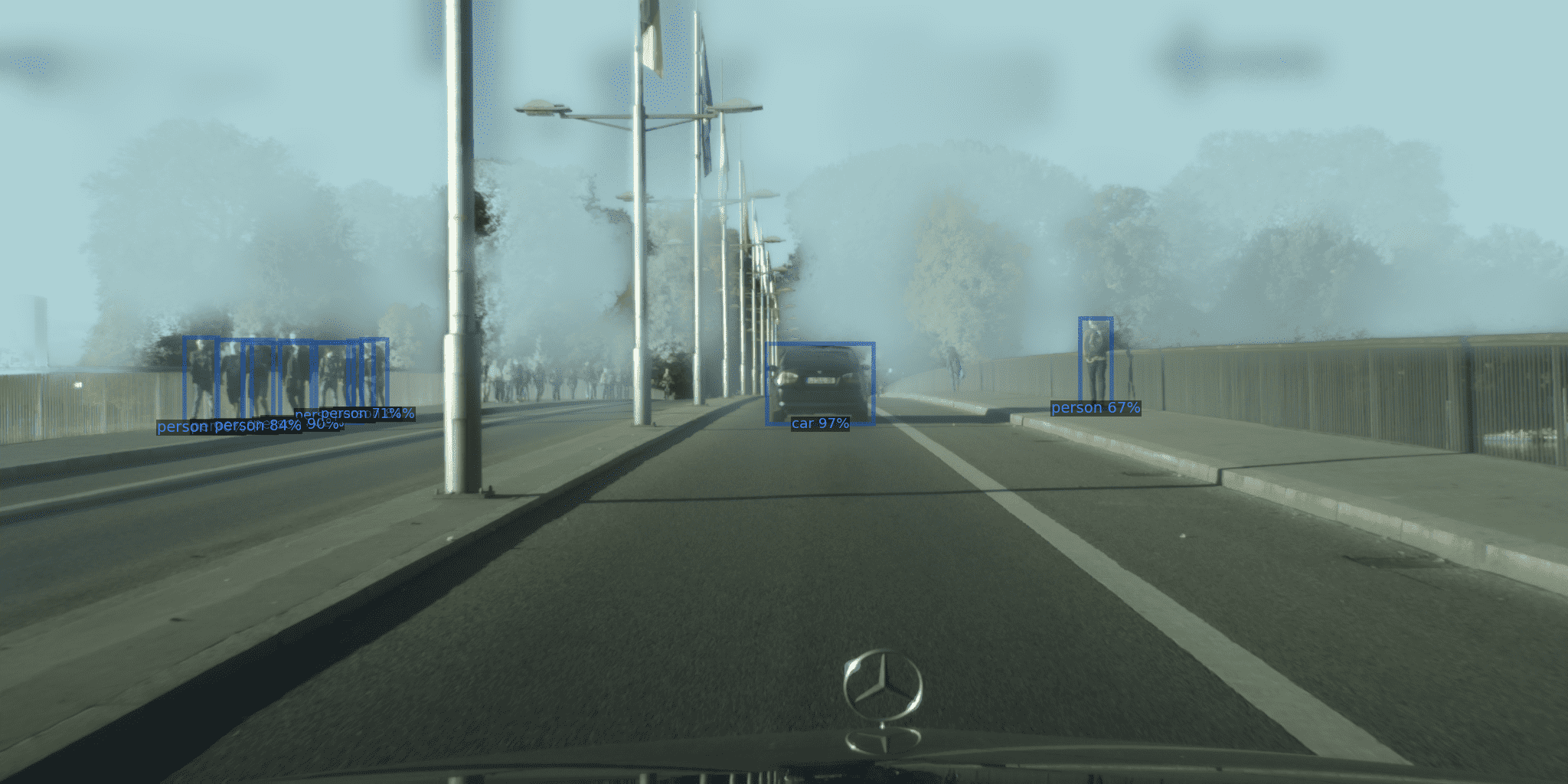} \\\vspace{2pt}
    \end{minipage}
    \begin{minipage}[b]{.235\linewidth}
        \centering
        \includegraphics[scale=.078]{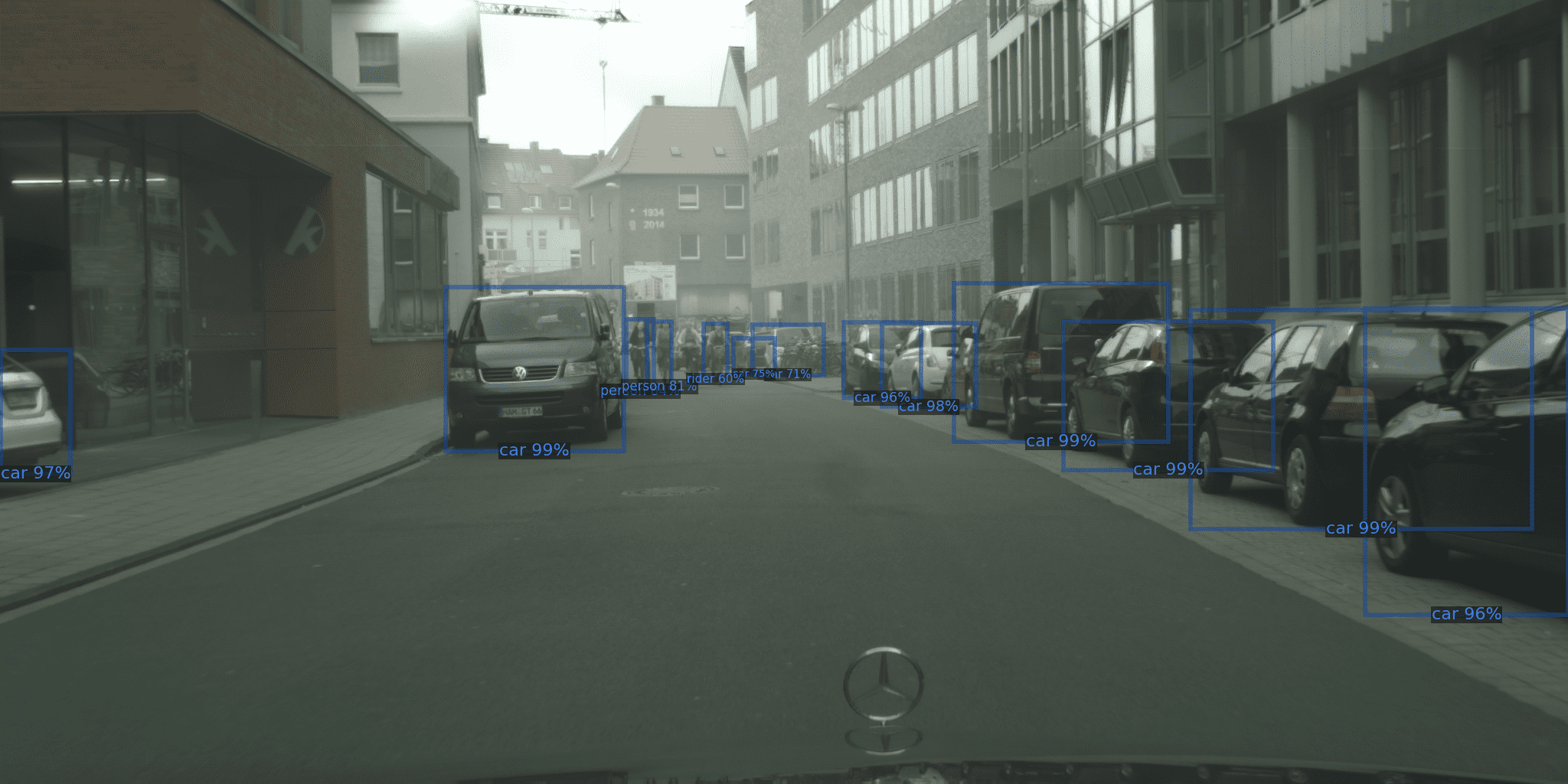} \\\vspace{2pt}
    \end{minipage}
    \begin{minipage}[b]{.235\linewidth}
        \centering  
        \includegraphics[scale=.078]{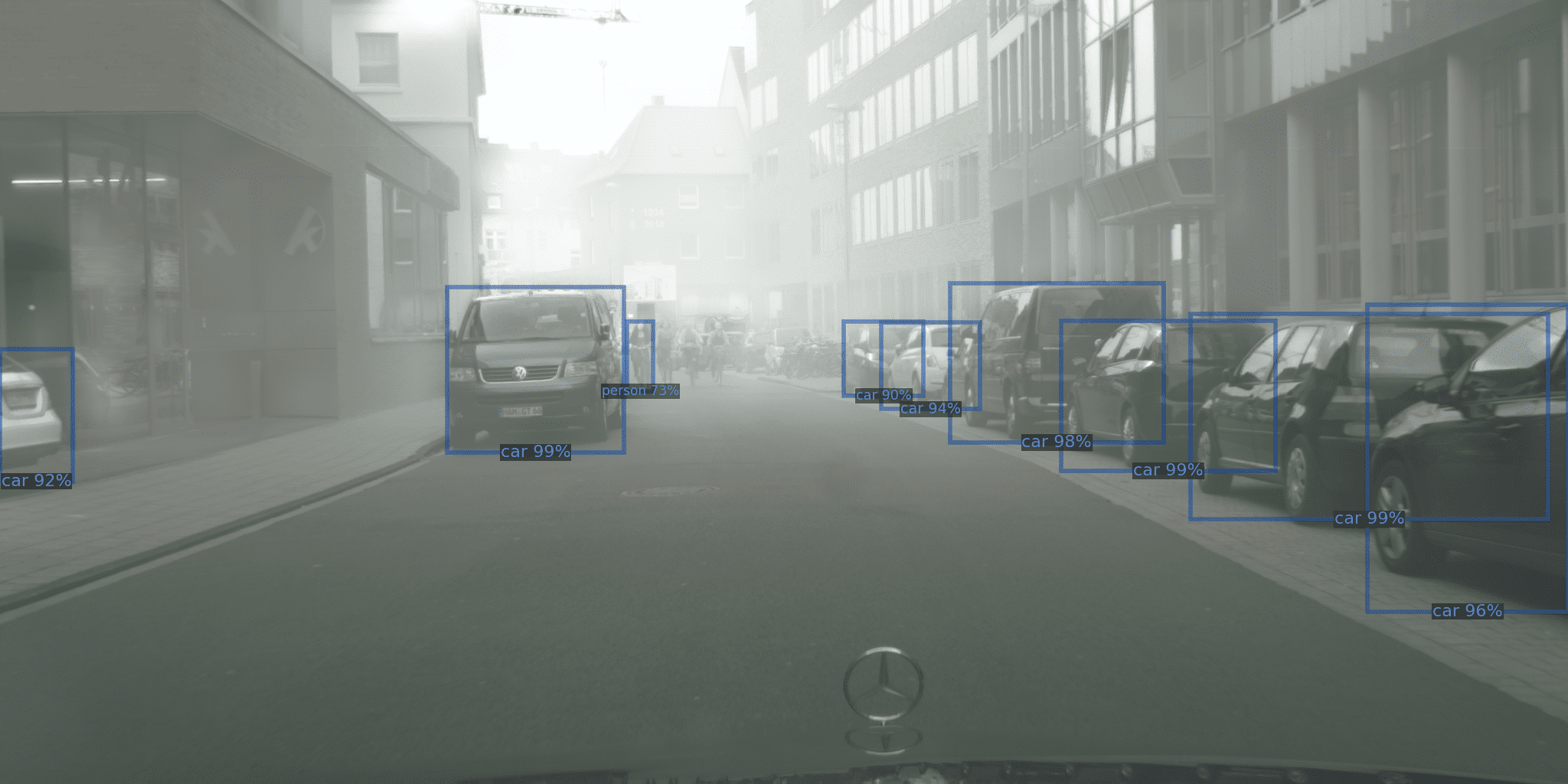} \\\vspace{2pt}
    \end{minipage}  
}
\\
\subfigure
{
    \rotatebox{90}{\scriptsize{~~~~~~~~~DT (4999)}}
    \begin{minipage}[b]{.235\linewidth}
        \centering
        \includegraphics[scale=.078]{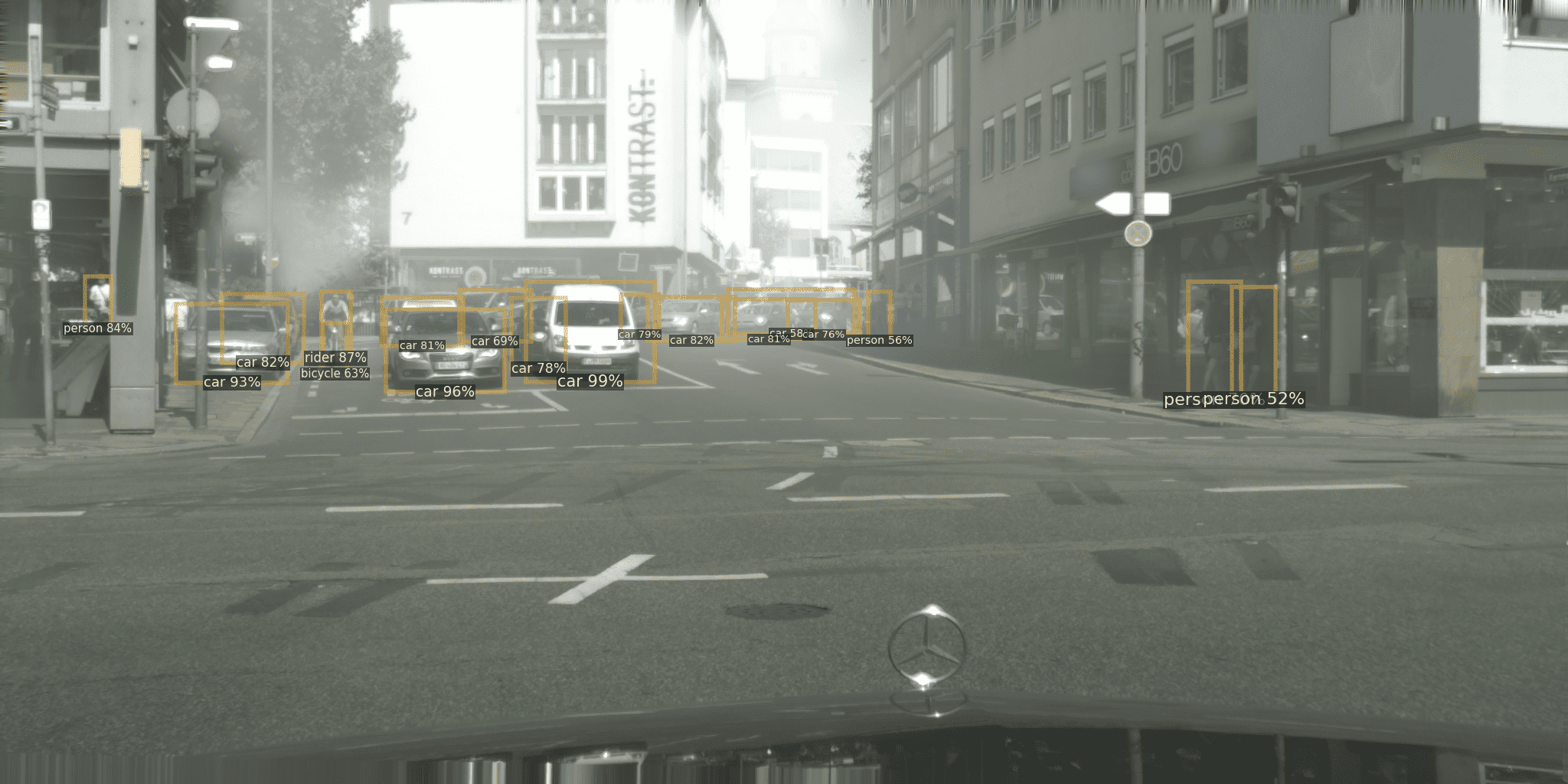}\vspace{2pt}
    \end{minipage}
    \begin{minipage}[b]{.235\linewidth}
        \centering
        \includegraphics[scale=.078]{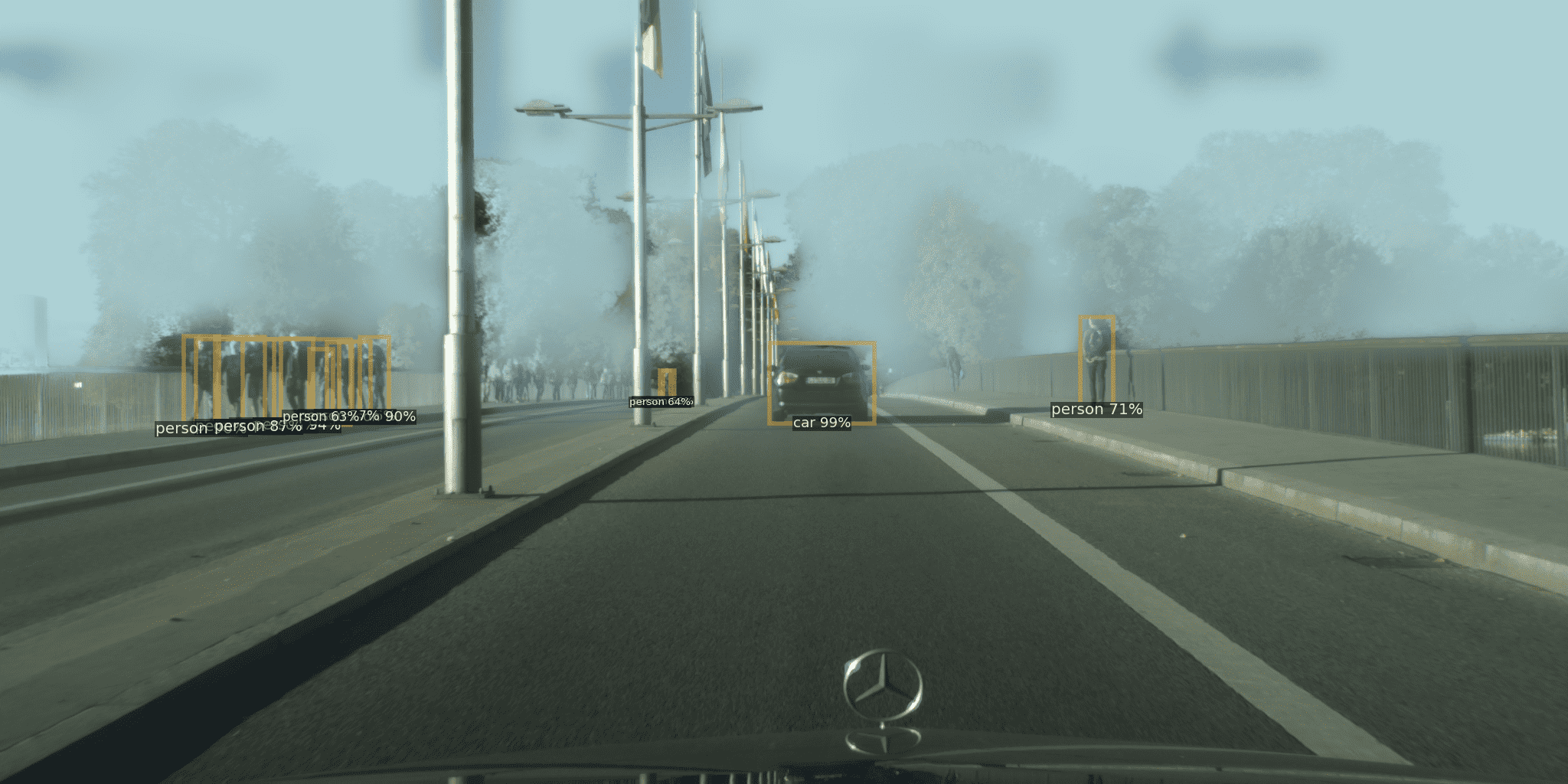} \\\vspace{2pt}
    \end{minipage}
    \begin{minipage}[b]{.235\linewidth}
        \centering
        \includegraphics[scale=.078]{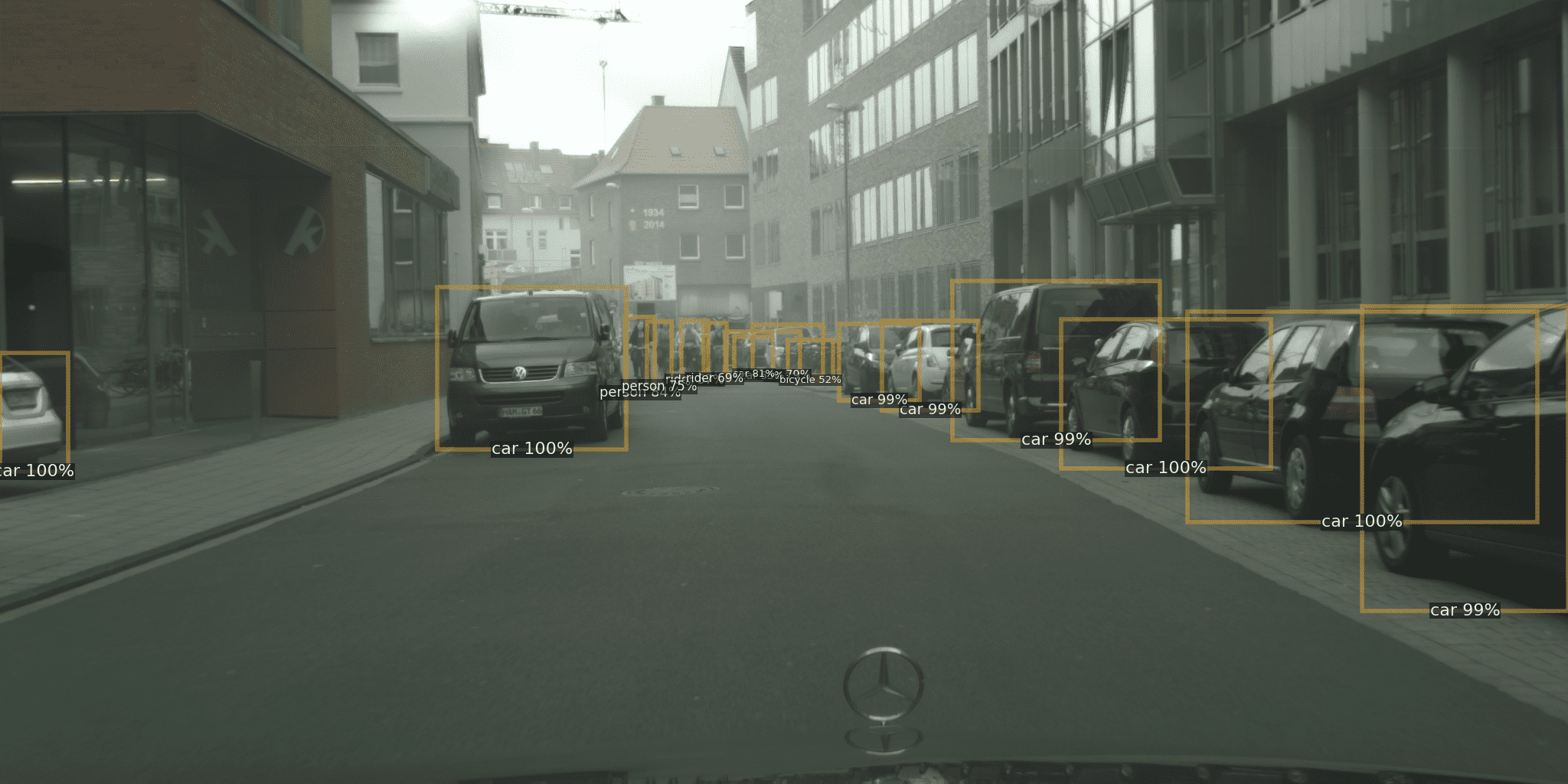} \\\vspace{2pt}
    \end{minipage}
    \begin{minipage}[b]{.235\linewidth}
        \centering  
        \includegraphics[scale=.078]{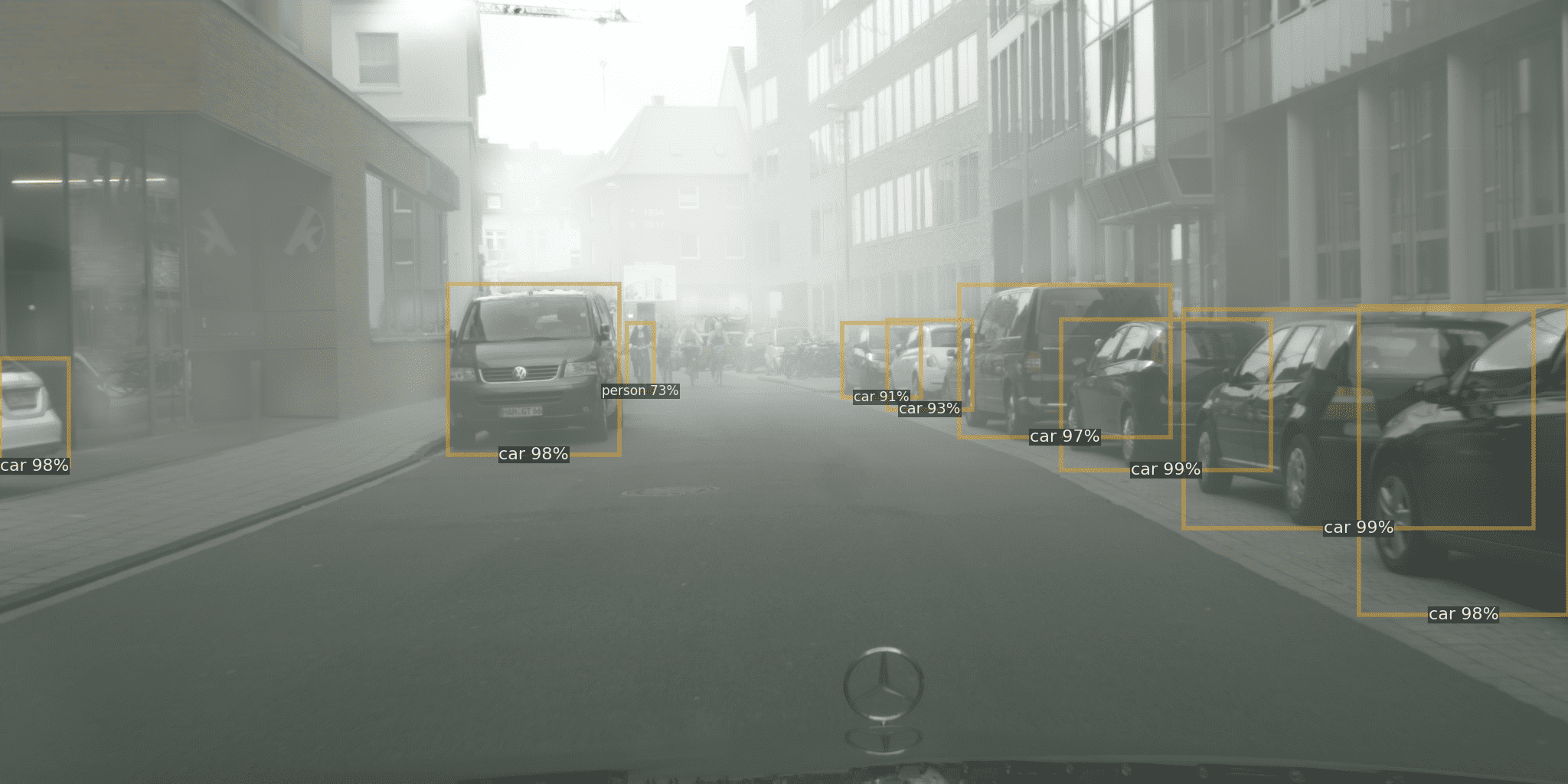} \\\vspace{2pt}
    \end{minipage}  
}
\\
\subfigure
{
    \rotatebox{90}{\scriptsize{~~~~~~Student (4999)}}
    \begin{minipage}[b]{.235\linewidth}
        \centering
        \includegraphics[scale=.078]{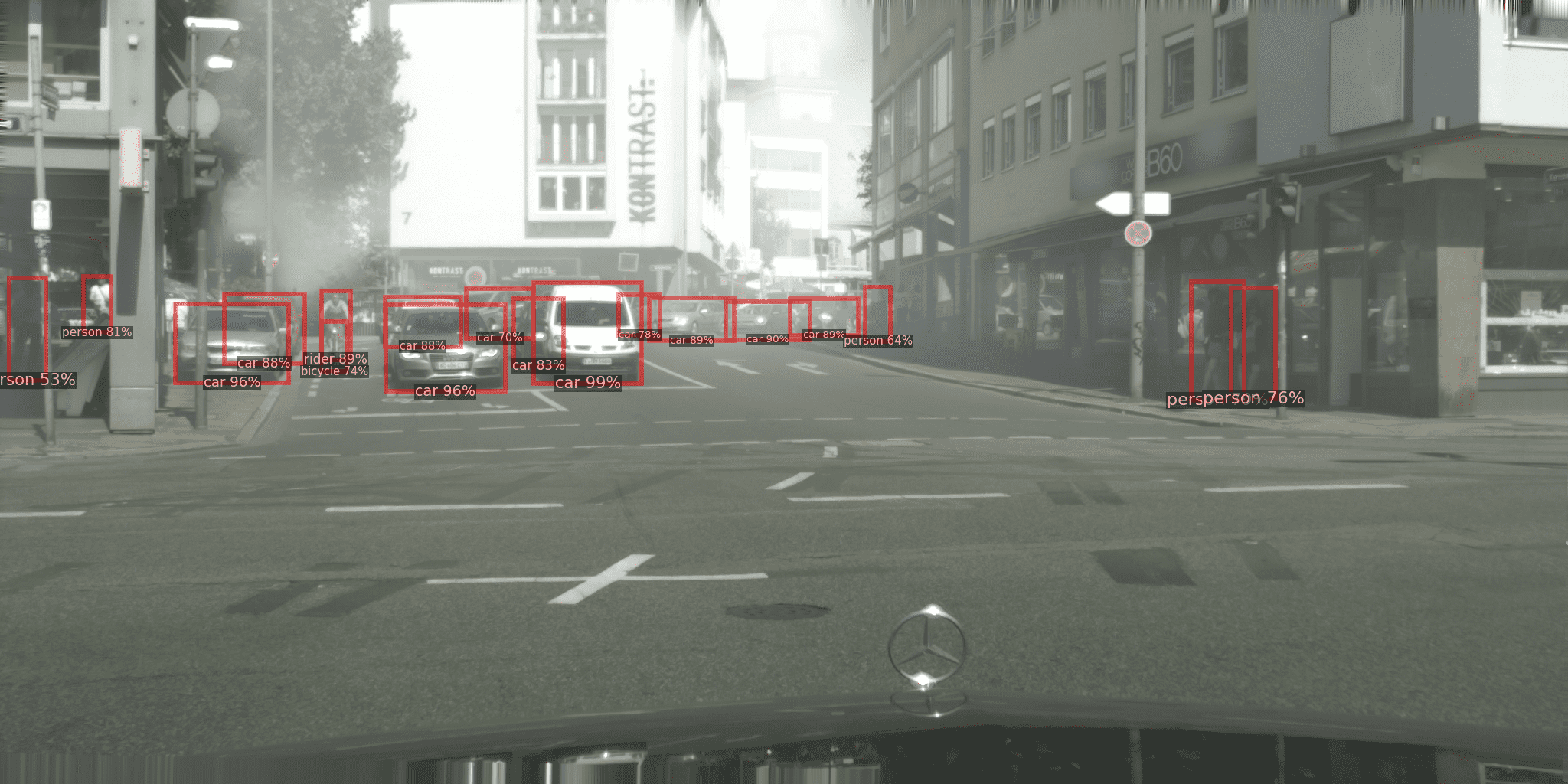}\vspace{2pt}
    \end{minipage}
    \begin{minipage}[b]{.235\linewidth}
        \centering
        \includegraphics[scale=.078]{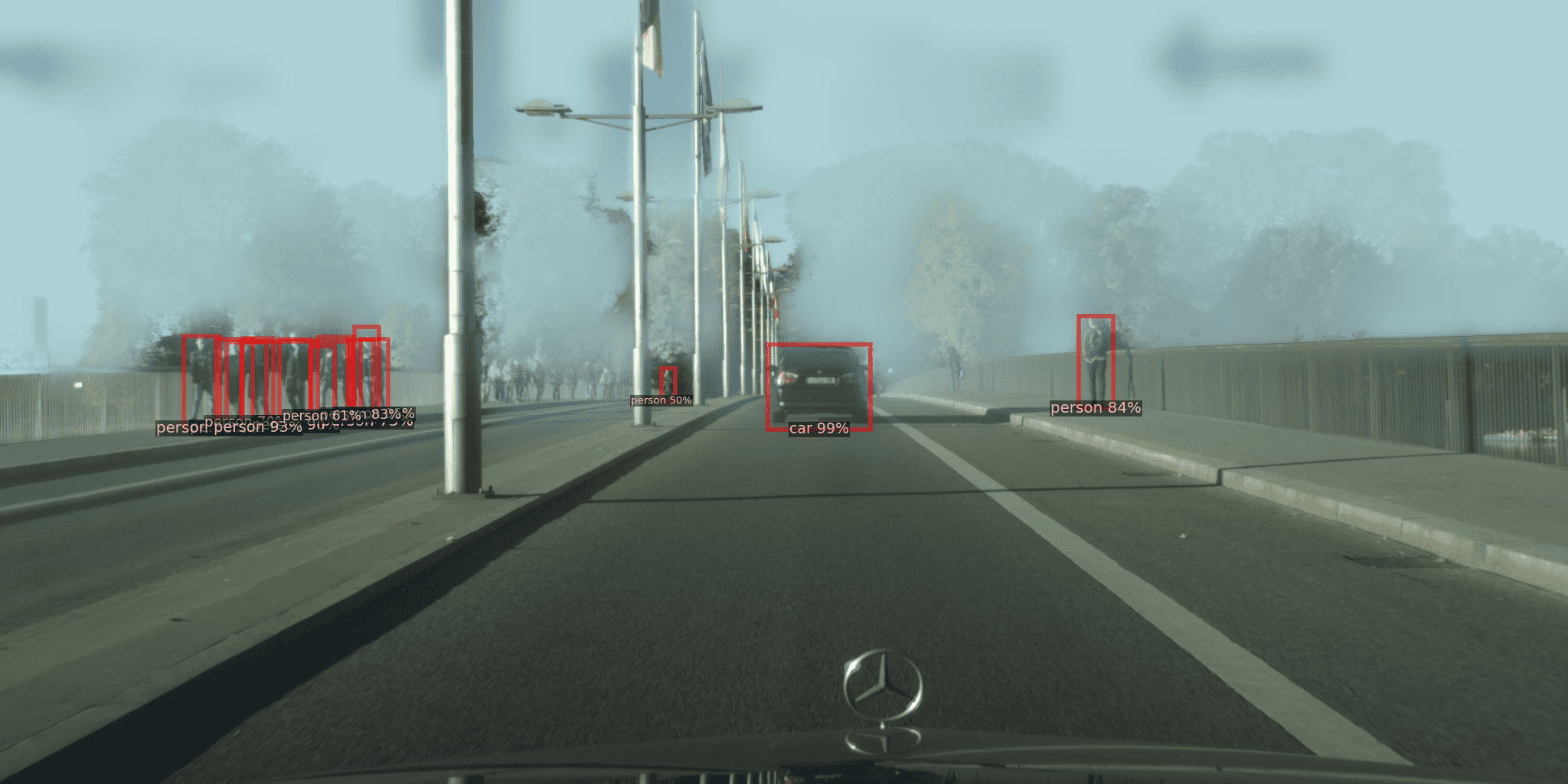} \\\vspace{2pt}
    \end{minipage}
    \begin{minipage}[b]{.235\linewidth}
        \centering
        \includegraphics[scale=.078]{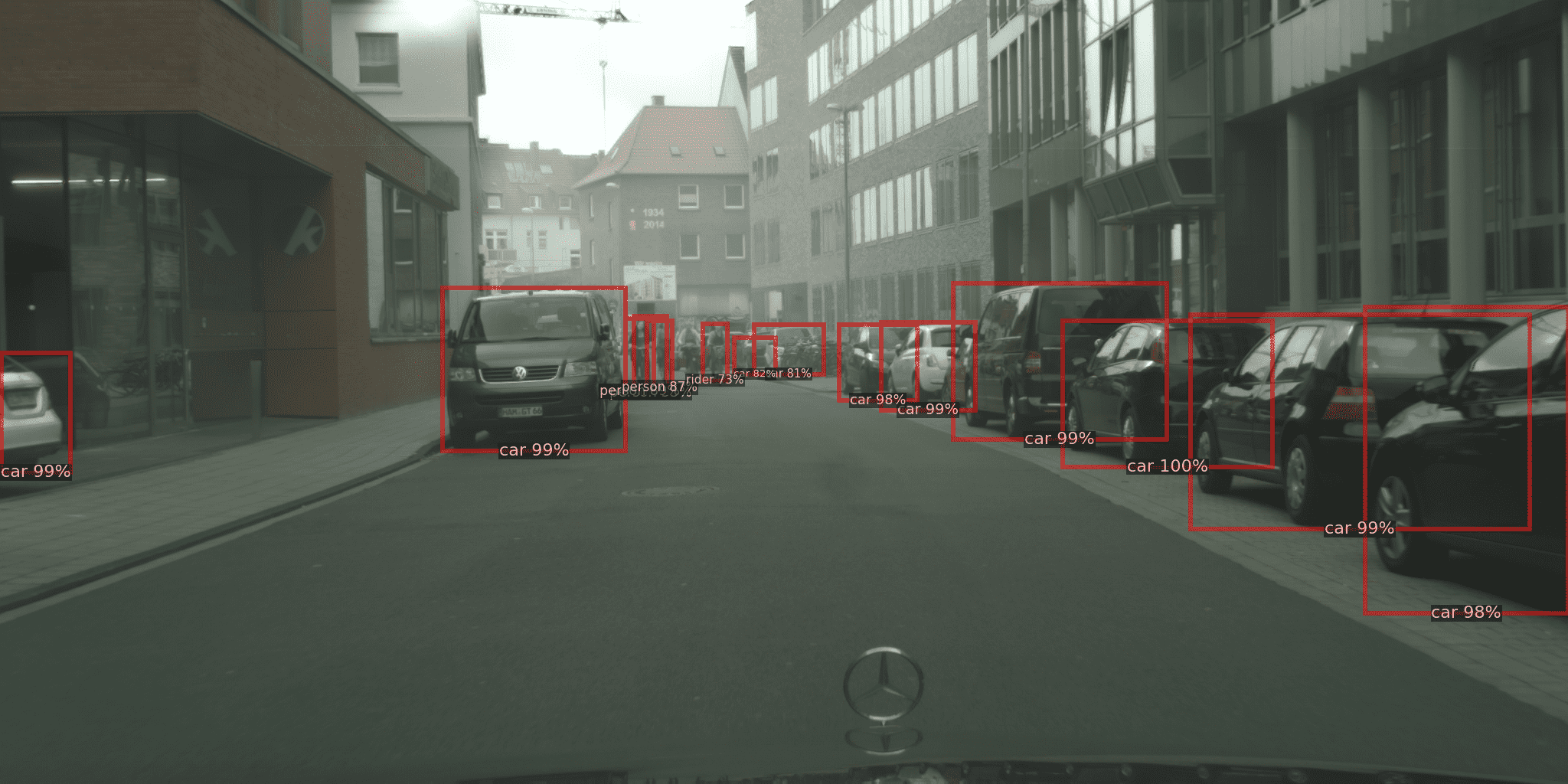} \\\vspace{2pt}
    \end{minipage}
    \begin{minipage}[b]{.235\linewidth}
        \centering  
        \includegraphics[scale=.078]{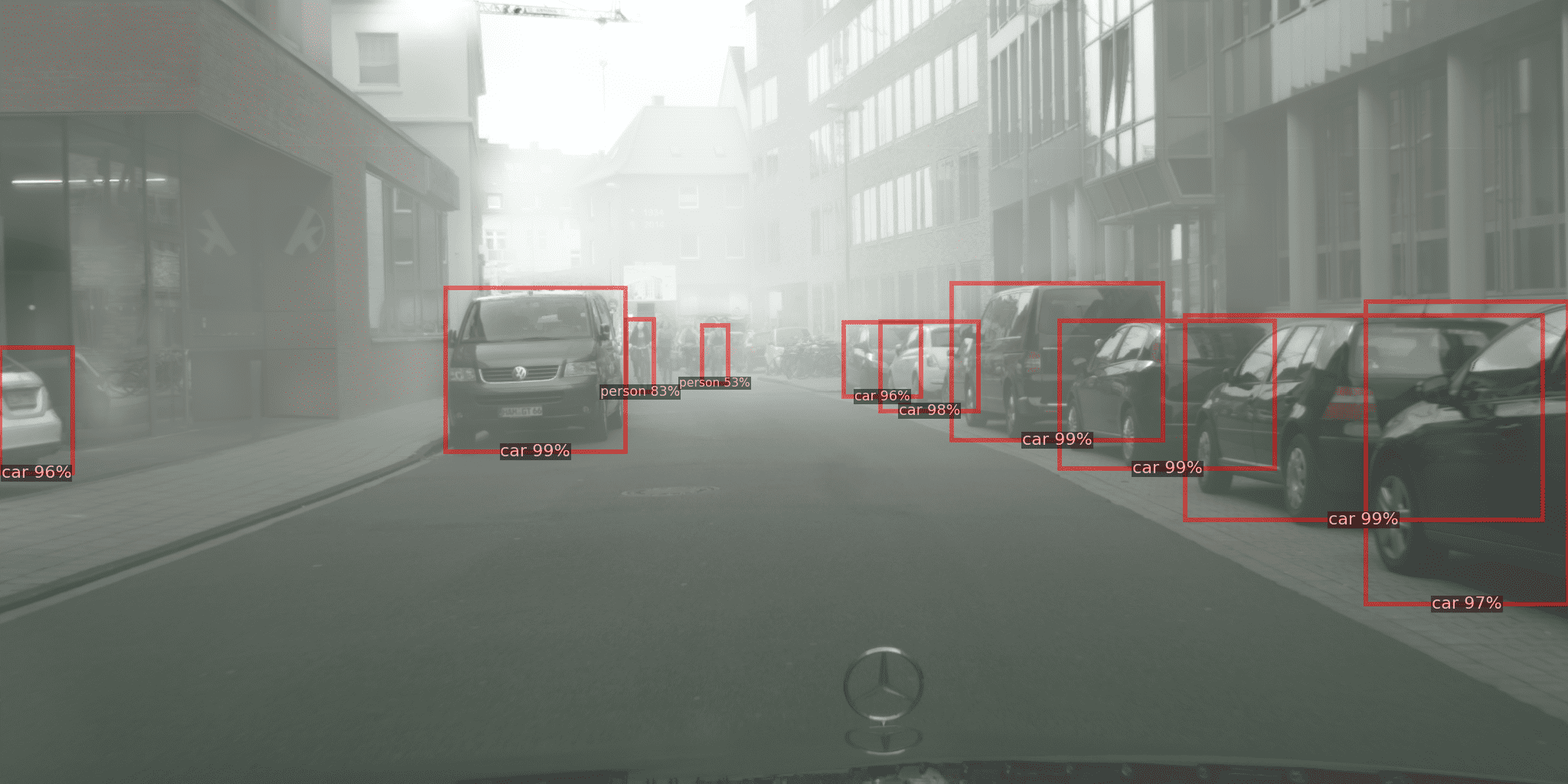} \\\vspace{2pt}
    \end{minipage}  
}
\\
\subfigure
{
    \rotatebox{90}{\scriptsize{~~~~~~~~~~~~Student (final)}}
    \begin{minipage}[b]{.235\linewidth}
        \centering
        \includegraphics[scale=.078]{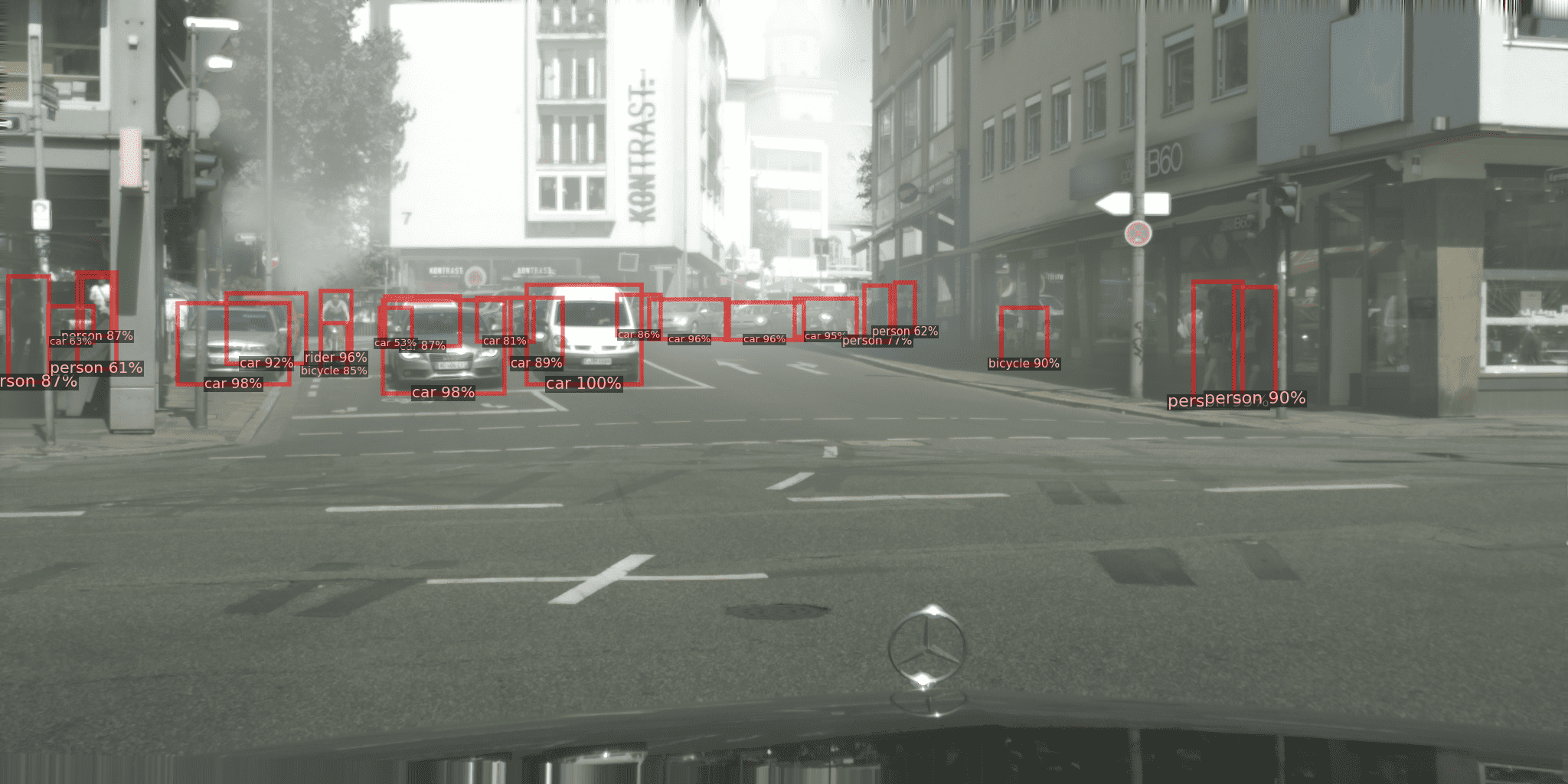} \\
        Foggy level = 0.01
    \end{minipage}
    \begin{minipage}[b]{.235\linewidth}
        \centering
        \includegraphics[scale=.078]{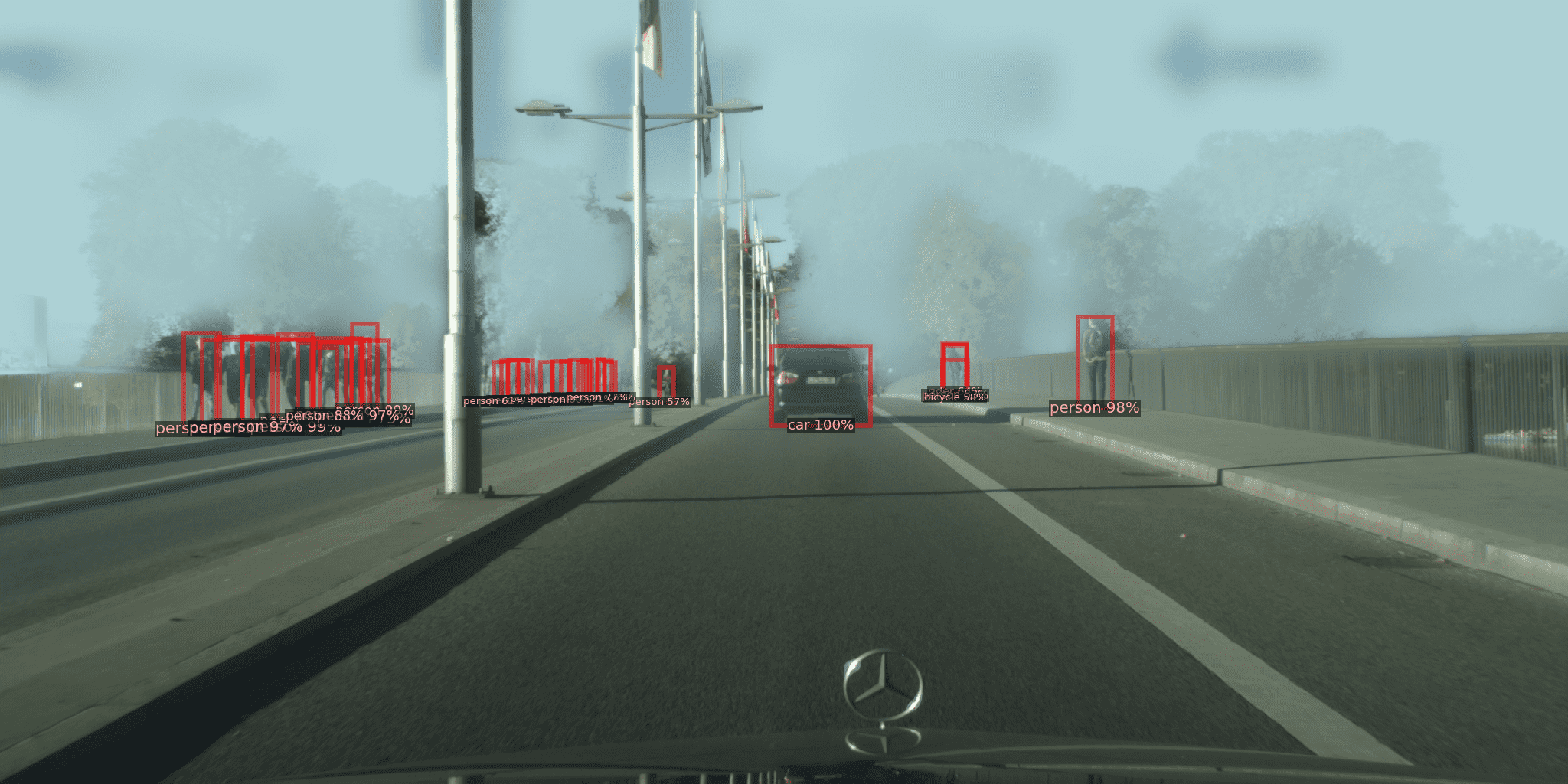} \\
        Foggy level = 0.01
    \end{minipage}
    \begin{minipage}[b]{.235\linewidth}
        \centering
        \includegraphics[scale=.078]{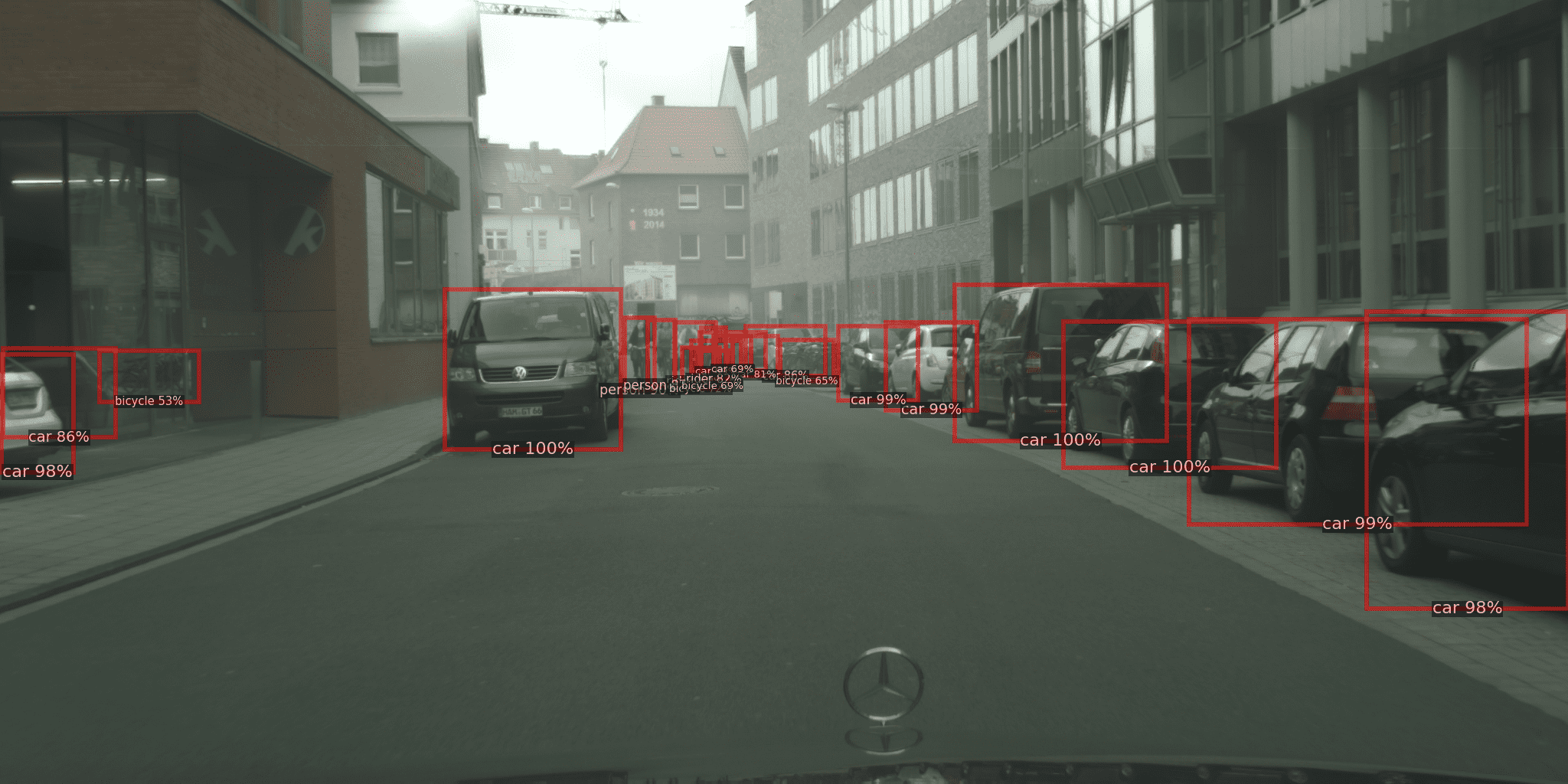} \\
        Foggy level = 0.005
    \end{minipage}
    \begin{minipage}[b]{.235\linewidth}
        \centering  
        \includegraphics[scale=.078]{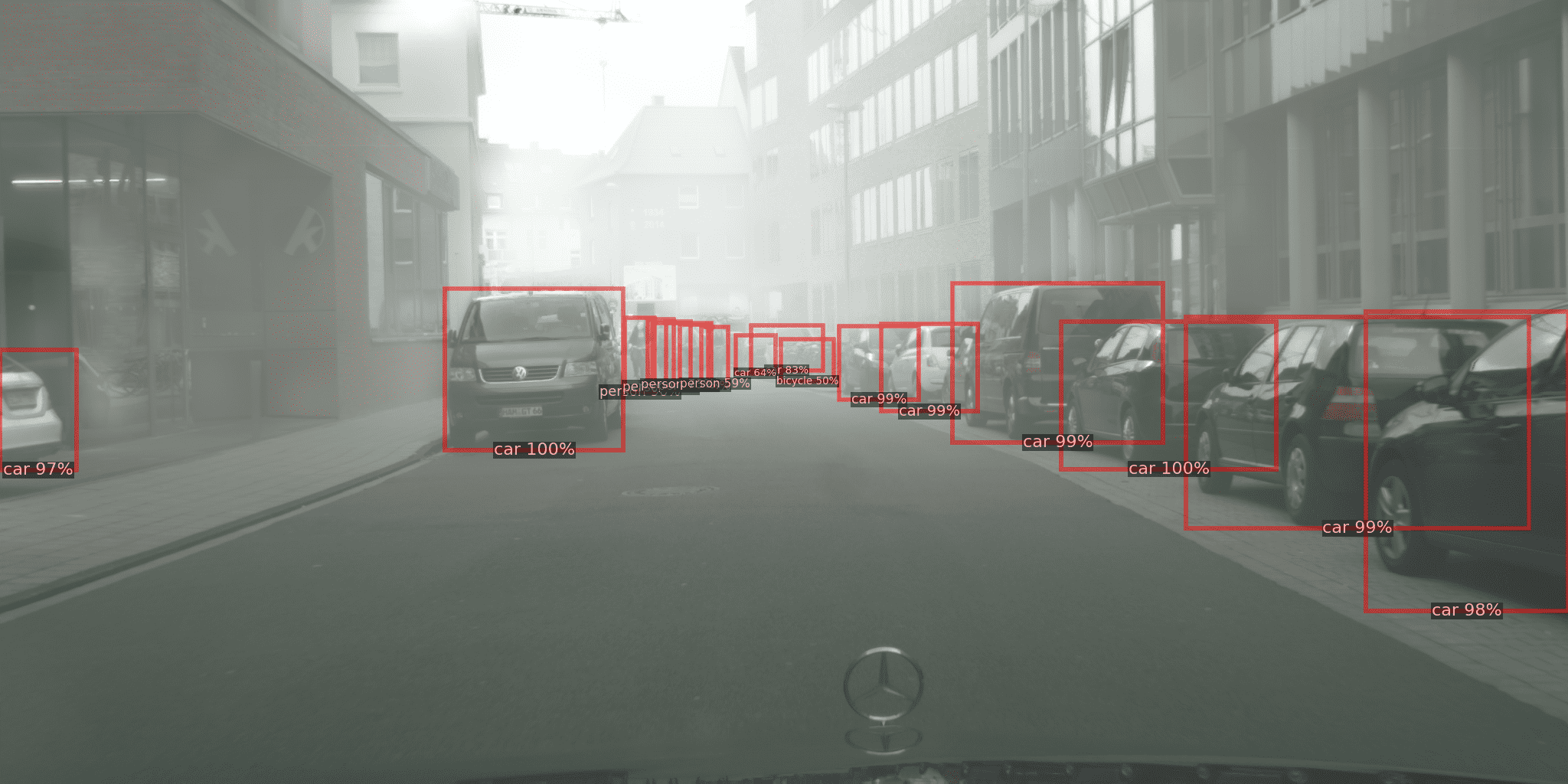} \\
        Foggy level = 0.02
    \end{minipage}  
}
\vspace{-0.4cm}
\caption{Detection results of different foggy-level images predicted by the dynamic teacher, static teacher, student models trained in different times. ``DT (4999)'', ``ST (4999)'', and ``Student (4999)'' represent the dynamic teacher, static teacher, and student model in the $4999$-th iteration, respectively. ``Student (final)'' represents the student model saved at the end of training.}
\vspace{-0.5cm}
\label{fig:big}
\end{figure*}

\subsection{Result Analysis}
\paragraph{Training Stability.}
The training curves of each model within our multi-teacher framework on the four benchmarks are shown in Figure \ref{fig:plot}. Compared with the training curves of the conventional MT framework (see Figure \ref{fig:slow}), the performance of the student, static teacher and dynamic teacher models is stably improved and gradually converges to a consistent point as the training progresses.
We can see that the training instability problem of conventional MT framework is effectively alleviated by our method. 

\paragraph{Visualization.}
We conduct an analysis by visualizing the detection results of the static teacher, dynamic teacher, and student models. This visualization is performed by inputting several images with varying foggy degrees from the Foggy Cityscape dataset~\cite{cityscapes_foggy}. The detection results of the three models for these images are shown in Figure \ref{fig:big}. It is evident that the two teacher models yield varying detection results for each image, implying the potential complementarity of their predictive results. This observation prompts us to make a consensus on the divergent predictions of the two teacher models to enhance the quality of pseudo labels. The effectiveness of the consensus mechanism is further proven by the detection results of the student model obtained at the final iteration, which has shown superior recall and accuracy compared to the student model at the intermediate ($4,999$-th) iteration.

\section{Conclusion}
In this paper, we present a simple yet novel \emph{Periodically Exchange Teacher-Student} method to tackle the training instability problem ignored by current MT-based SFOD methods. Our method employs a static teacher model, a dynamic teacher model, and a student model. At the end of each training period, we exchange the weights between the static teacher and student models. Within each period, the static teacher maintains its weights, while the student model is trained using pseudo labels generated by both teachers. Meanwhile, the dynamic teacher is continually updated using the EMA of the student model per iteration throughout the whole training phase. The extensive experimental results demonstrate the effectiveness of our method. Our method provides a new insight for MT-based self-training methods.

\section*{Acknowledgements}
This work was supported by the Fujian Provincial Natural Science Foundation (No. 2022J05135), the University-Industry Project of Fujian Provincial Department of Science and Technology (No. 2020H6005), and the National Natural Science Foundation of China (No. U21A20471).

{\small
\bibliographystyle{ieee_fullname}
\bibliography{egpaper_final}
}

\end{document}